\documentclass[hyperref={colorlinks,citecolor=red,urlcolor=black,bookmarks=false,hypertexnames=true}]{article}

\usepackage{arxiv}

\usepackage[utf8]{inputenc} 
\usepackage[T1]{fontenc}    
\usepackage{url}            
\usepackage{booktabs}       
\usepackage{amsfonts}       
\usepackage{nicefrac}       
\usepackage{microtype}      
\usepackage{lipsum}         
\usepackage{graphicx}
\usepackage{subcaption}
\usepackage{doi}

\usepackage[english]{babel}

\usepackage{colortbl}
\usepackage{listings}
\usepackage{bm}
\usepackage{physics}

\usepackage{amsmath,accents}
\usepackage{cleveref}       

\usepackage[ruled,vlined]{algorithm2e}

\usepackage{appendix}
\usepackage{cite}

\usepackage{multirow}
\usepackage{array}

\usepackage{chngcntr}
\usepackage{epigraph}

\usepackage{gensymb}

\usepackage{import}
\usepackage{xifthen}
\usepackage{pdfpages}
\usepackage{transparent}

\SetKwComment{Comment}{/* }{ */}

\newcommand{\ttl}{%
		Physics-Informed Neural Systems for\\ the Simulation of EUV Electromagnetic Wave\\ Diffraction from a Lithography Mask
}

\title{\ttl}

\date{\today}

\author{ \href{https://orcid.org/0000-0002-4930-1846}{\includegraphics[scale=0.06]{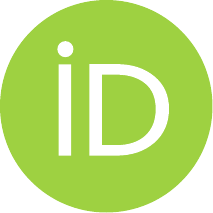}\hspace{1mm}Vasiliy A. Es'kin}\thanks{Corresponding author: Vasiliy Alekseevich Es’kin (\href{mailto:vasiliy.eskin@gmail.com}{vasiliy.eskin@gmail.com})}, {Egor V. Ivanov} \\
	Department of Radiophysics, University of Nizhny Novgorod\\
	 23 Gagarin Ave., Nizhny Novgorod 603022, Russia\\
	\href{mailto:vasiliy.eskin@gmail.com}{\texttt{vasiliy.eskin@gmail.com}}, {\texttt{iev90078@gmail.com}}
}

\renewcommand{\headeright}{}
\renewcommand{\undertitle}{}
\renewcommand{\shorttitle}{A preprint}

\renewcommand{\vec}{\bf}

\newcommand{\rot}{{\rm rot}}
\renewcommand{\div}{{\rm div}}
\newcommand{\eps}{\varepsilon}

\usepackage{tikz}
\newcommand\tikznode[3][]%
{\tikz[remember picture,baseline=(#2.base)]
	\node[minimum size=0pt,inner sep=0pt,#1](#2){#3};%
}

\hypersetup{
pdftitle={Physics-Informed Neural Systems for the Simulation of EUV Electromagnetic Wave Diffraction from a Lithography Mask},
pdfsubject={math.na, cs.AI, physics.comp-ph},
pdfauthor={Vasiliy A. Es'kin},
pdfkeywords={Deep Learning, Physics-informed Neural Networks, Neural Operator, Predictive modeling, Computational physics, Extreme Ultraviolet Lithography}
}

\begin{document}
\maketitle

\begin{abstract}
	Physics-informed neural networks (PINNs) and neural operators (NOs) for solving the problem of diffraction of Extreme Ultraviolet (EUV) electromagnetic waves from contemporary lithography masks are presented. A novel hybrid Waveguide Neural Operator (WGNO) is introduced, based on a waveguide method with its most computationally expensive components replaced by a neural network. To evaluate performance, the accuracy and inference time of PINNs and NOs are compared against modern numerical solvers for a series of problems with known exact solutions. The emphasis is placed on investigation of solution accuracy by considered artificial neural systems for 13.5 nm and 11.2 nm wavelengths. Numerical experiments on realistic 2D and 3D masks demonstrate that PINNs and neural operators achieve competitive accuracy and significantly reduced prediction times, with the proposed WGNO architecture reaching state-of-the-art performance. The presented neural operator has pronounced generalizing properties, meaning that for unseen problem parameters it delivers a solution accuracy close to that for parameters seen in the training dataset.  These results provide a highly efficient solution for accelerating the design and optimization workflows of next-generation lithography masks.

\end{abstract}


\section{Introduction}
Extreme ultraviolet lithography is a critical component in the development of semiconductor manufacturing processes. This technology enables the creation of smaller and more advanced semiconductor chips by utilizing shorter wavelengths of light to etch intricate designs onto silicon wafers. Through this method, manufacturers are able to continue scaling down the size of transistors in accordance with Moore's Law.

\begin{figure}[ht!]
	\centering
	\includegraphics[width=0.8\linewidth]{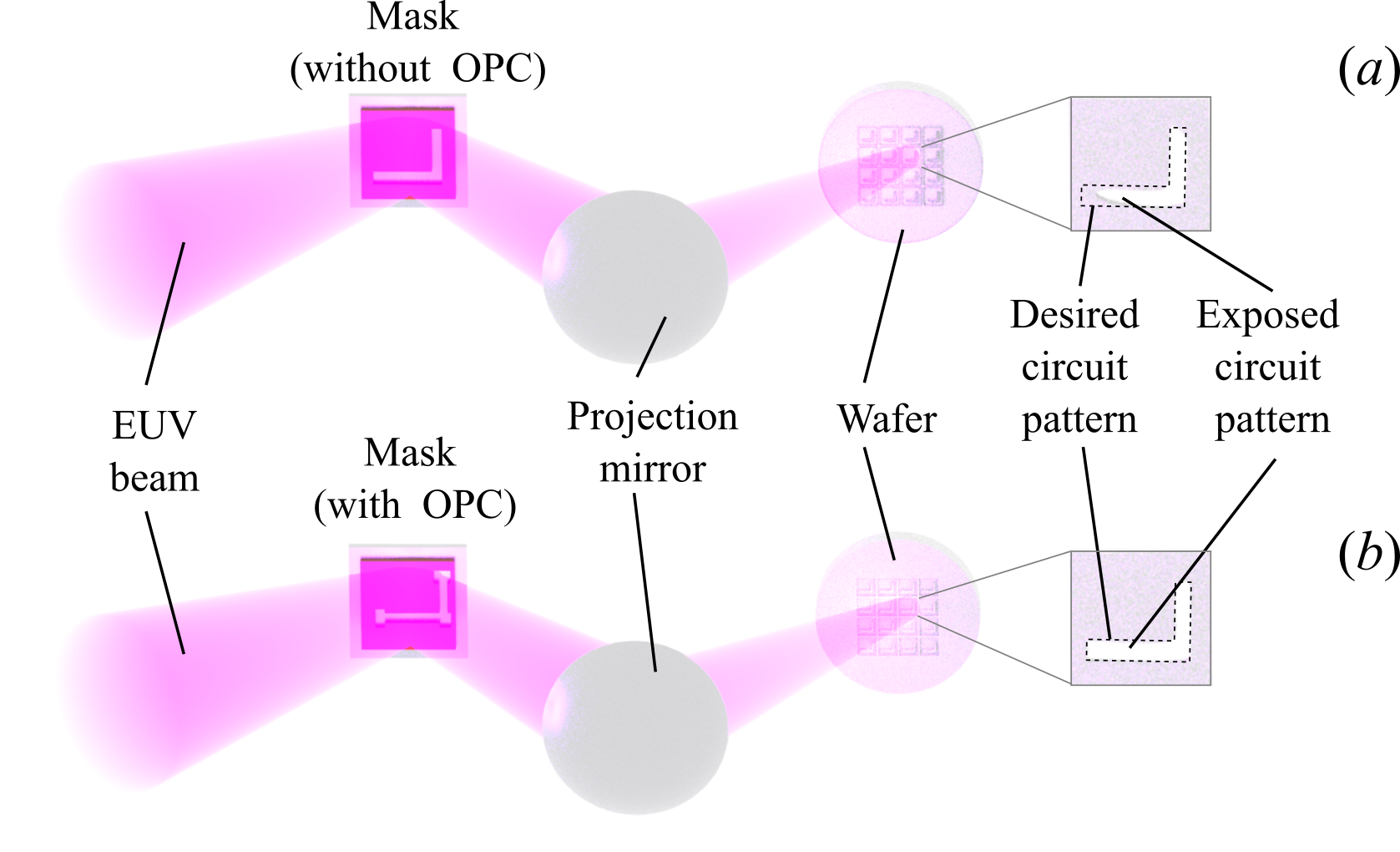}
	\caption{EUV lithography with (a) and without (b) optical proximity correction of mask.}
	\label{fig1}
\end{figure}

Integrated circuits are made using EUV lithography, a process analogous to photographic printing, in which the patterns that will become layers of an integrated circuit are exposed on a semiconductor wafer, one layer at a time. The light pattern on the wafer is formed due to the reflection of EUV electromagnetic beam from the mask (see Figure~\ref{fig1}). Modern integrated circuits require more than 80 different masks in their production. Due to diffraction and interference phenomena, the patterns on the mask do not match the light pattern on the wafer (see Figure~\ref{fig1} (a)). To obtain the desired pattern on the wafer, a multi-stage technology of optical proximity correction (OPC) of mask is used (Figure~\ref{fig1} (b)). One of the stages of optical approximation correction is calculating electromagnetic fields in the area of the mask location. The diffracted electromagnetic waves can be calculated rigorously by using electromagnetic (EM) simulators. However, these calculations require massive computational resources.

To accelerate the EM simulations, various approximation models have been proposed, including the ``domain decomposition method'' and the ``M3D filter''~\cite{Tanabe24}. These models break down a mask pattern into two-, one-, and zero-dimensional patterns. In these models, the electromagnetic (EM) field of a mask pattern is calculated by combining the EM fields from 2D, 1D, and 0D components. These models are currently employed in numerous EUV lithography simulation tools~\cite{Tanabe24}. However, these models are still very resource-intensive and do not take into account the nonlocality of electromagnetic interaction.

Recently, many attempts have been made to simulate the 3D effects of masks using deep neural networks (DNNs) such as convolutional neural networks (CNNs)~\cite{tanabe2024accelerating}, generative adversarial network (GAN) or U-Net~\cite{Medvedev2023,Medvedev2024b}. However, these neural network approaches are based on supervised learning, require fairly extensive datasets, have a significant training time, and often do not demonstrate the necessary accuracy of the solution and the degree of its generalization.

In this paper, we present physics-informed neural networks~\cite{Raissi2019,Eskin2024,ESKIN2025114085} and neural operators~\cite{lu2021learning,li2021fourierneuraloperatorparametric} for solving the problem of diffraction of extreme ultraviolet electromagnetic waves from 3D masks. The training of these neural systems is performed in an unsupervised manner, leveraging the governing physical equations directly in the training process. We evaluate the performance and accuracy of the neural network systems on benchmark problems and on realistic 2D and 3D mask models for current industrial lithography systems (13.5 nm)~\cite{yakshin2000determination} and of promising ones (11.2 nm)~\cite{Chkhalo2024,Abramov2025}.

The paper is structured as follows. In section 2, physical  problem is formulated and basic equations are given. Section 3 includes the description of solution methods, which are the finite element method, waveguide method, methods based on physics-informed neural networks and neural operators. Section 4 presents the evaluation of these methods on problems with exact solutions, as well as numerical experiments applying these methods to simulate electromagnetic wave diffraction from realistic masks. Finally, in Section 5 concluding remarks are given.

\section{Formulation of the problem and basic equations}

\begin{figure}[ht!]
	\centering
	\includegraphics[width=0.4\linewidth]{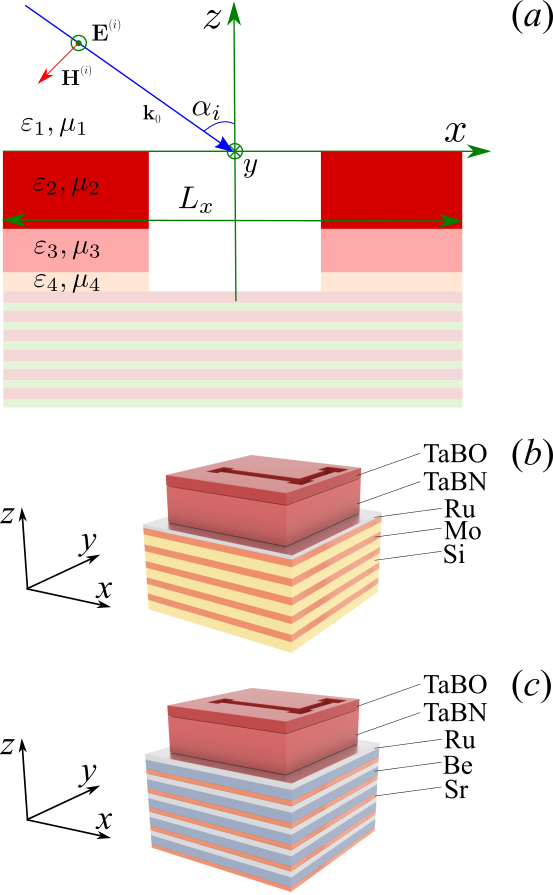}
	\caption{Geometry of the problem}
	\label{fig2}
\end{figure}

Consider a layered mask of thickness $D$, consisting of $J$ layers, located in free space and enclosed in an interval $[-D,0]$ along the $z$-axis of a Cartesian coordinate system ($x,y,z$), as shown in Fig.~\ref{fig2} (a). The mask structure is periodic along the $x$ and $y$ axes with periods $L_x$ and $L_y$, respectively. Each $j$th layer of the mask is filled with a medium, which is uniform in the $z$ direction and has a dielectric permittivity of the $j$-th layer is $\eps_j = \eps_j(x,y)$. The dielectric permittivities of the media were obtained from experimental data published in~\cite{HENKE1993181,CenterXRayOpt}.

We consider the diffraction of an electromagnetic monochromatic plane wave with angular frequency $\omega$ from a mask (see Fig.~\ref{fig2}(a)). The  magnetic fields in the incident wave are given, with $\exp(i\omega t)$ time dependence dropped, by ${\vec H}^{(i)}={\vec H}_0\exp[-i \left(k_{0;x} x + k_{0;y} y - k_{0;z} z\right)]$, where ${\vec H}_0$ is the magnetic field amplitude, $k_{0;x}$, $k_{0;y}$, and $k_{0;z}$ are the components of the wave vector ${\vec k}_0$ in free space ($k_0 = \left(k_{0;x}^2 + k_{0;y} ^ 2 + k_{0;z}^2\right)^{1/2}$, $k_0 = \omega /c$, where $c$ is the speed of light in free space), and the superscript $(i)$ denotes the incident wave. To simplify further consideration of the problem, we assume that the field of incident wave is periodic along the $x$ and $y$ axes with periods $L_x$ and $L_y$, respectively.

We assume that the media of the mask are time-independent and nonmagnetic (the magnetic permeability $\mu=1$). Maxwell equations for a monochromatic field in the Gaussian system of units are written as follows:
\begin{align}
	& {\rot \vec{E}} = -i k_0 \vec{H}, \notag\\
	& {\rot \vec{H}} = i k_0 \eps \vec{E}, \notag \\
	& \div \left({\eps \vec{E}}\right) = 0, \notag \\
	& \div {\vec H} = 0.\label{eq1}
\end{align}
It can be shown that the vector components $\vec{E}$ and $\vec{H}$ of electromagnetic field in a homogeneous media along the $z$ axis within each $j$th layer are expressed in terms of magnetic components $H_x$ and $H_y$ which, in turn, satisfy the following system of equations:
\begin{align}
	&\Delta H_x + k_0^2 \varepsilon_j H_x + \frac{1}{\varepsilon_j} \frac{\partial \varepsilon_j}{\partial y} \left( \frac{\partial H_{y}}{\partial x} - \frac{\partial H_{x}}{\partial y} \right){=} 0,\notag \\
	&\Delta H_y + k_0^2 \varepsilon_j H_y +  \frac{1}{\varepsilon_j}\frac{\partial \varepsilon_j}{\partial x} \left( \frac{\partial H_{x}}{\partial y} - \frac{\partial H_{y}}{\partial x} \right){=} 0. \label{eq2}
\end{align}

The problem is solved within a computational domain subject to boundary conditions. Periodicity conditions are satisfied on the lateral boundaries
\begin{align}
	& {\vec{H}}\left(-L_x/2, z\right) = {\vec{H}}\left(L_x/2, z\right),\notag \\
	&{\vec{H}}\left(-L_y/2, z\right) = {\vec{H}}\left(L_y/2, z\right).\label{eq3}
\end{align}
At the top ($z=z_{\max}$) and bottom ($z=z_{\min}$) of the domain, Sommerfeld radiation conditions are applied to ensure that scattered waves propagate outwards without unphysical reflections. These are formulated for every spatial Fourier harmonic of the field as follow:
\begin{equation}\label{eq4}
	\int\limits_{-L_x/2}^{L_x/2} \int\limits_{-L_y/2}^{L_y/2} \left(\frac{\partial {{\vec H}_\perp}}{\partial z} \pm i {k}_{z;mn} {{\vec H}_\perp} \right) e^{i (\kappa_x m x + \kappa_y n y)} dx dy = 0,
\end{equation}
where ``$+$'' and ``$-$'' correspond to $z=z_{\max}$ and  $z=z_{\min}$, respectively, ${k}_{z;mn} = (k_0^2 - \kappa_x^2 m^2 - \kappa_y^2 n^2)^{1/2}$ (branch ${\rm Re}\,({k}_{z;mn}) \geq 0$ and ${\rm Im}\,({k}_{z;mn}) \leq 0$ is taken), $\kappa_x = 2 \pi / L_x$, $\kappa_y = 2 \pi / L_y$, $m$ and $n \in \mathbb{Z} $.

\section{Solution Methods}

\subsection{Finite Element Method (FEM)}
The Finite Element Method is a powerful and general numerical technique for solving partial differential equations. The core principle of FEM involves discretizing the computational domain into a mesh of smaller, simpler subdomains called finite elements. Within each element, the unknown physical field is approximated by a set of basis functions. Assembling the equations for all elements leads to a large global system of linear equations, which is then solved to obtain the approximate solution across the entire domain. While robust, FEM can be computationally expensive for problems like EUV diffraction, which require very fine meshes to resolve the small wavelength features. In our comparative studies, we utilize the open-source solver FreeFem++~\cite{hecht2012new}.

\subsection{Waveguide Method}

The Waveguide (WG) method serves as our high-fidelity reference solver~\cite{tanabe2024accelerating,yuan1991modeling,lucas1996efficient}. It leverages the layered structure of the mask. Within each layer, the field is represented as a superposition of waveguide modes. The solution we will be found in the following form~\cite{yuan1991modeling,lucas1996efficient,Tanabe24}
\begin{align}\label{eq5}
	& H_x (x,y,z) = Z(z) h_x(x,y), \notag\\
	& H_y (x,y,z) = Z(z) h_y(x,y).
\end{align}
We assume that the $z$ dependence can be described by
\begin{align}
	& Z(z) = \exp\left(- i k_z z\right).
\end{align}
The equations for the components $h_x$ and $h_y$ are
\begin{align}\label{eq7}
	& \Delta_\perp h_x + k_0^2 \eps h_x - k_z^2 h_x +  \frac{1}{\eps} \frac{\partial \eps}{\partial y} \left( \frac{\partial h_{y}}{\partial x} - \frac{\partial h_{x}}{\partial y} \right) = 0, \notag\\
	& \Delta_\perp h_y + k_0^2 \eps h_y - k_z^2 h_y +  \frac{1}{\eps}\frac{\partial \eps}{\partial x} \left( \frac{\partial h_{x}}{\partial y} - \frac{\partial h_{y}}{\partial x} \right) = 0.
\end{align}
In this subsection the subscript of layer $j$ is dropped for brevity. The functions $h_x$ and $h_y$ will be found in the form of a Fourier series:
\begin{align}
	& h_x(x,y) = \sum\limits_{m,n = -\infty}^{\infty} B_{mn}\psi_{mn},\notag\\
	& h_y(x,y) = \sum\limits_{m,n = -\infty}^{\infty} C_{mn}\psi_{mn},\label{eq8}\\
	& \psi_{mn} = \exp\left(- i \kappa_x m x - i \kappa_y n y\right). \notag
\end{align}
From the system of Equations (\ref{eq7}) we have the following equations
\begin{align}
	-\sum\limits_{m,n} (\kappa_x^2 m^2 + \kappa_y^2 n^2) B_{mn}\psi_{mn} & + k_0^2 \sum\limits_{l,p} \eps_{lp}\psi_{lp} \sum\limits_{m,n} B_{mn}\psi_{mn}\notag\\
	& - \sum\limits_{l,p} \hat{\eps}^{(y)}_{lp}\psi_{lp} \left( \sum\limits_{m,n} i \kappa_x m C_{mn}\psi_{mn} - \sum\limits_{m,n} i \kappa_y n B_{mn}\psi_{mn} \right) = k_z^2 \sum\limits_{m,n} B_{mn}\psi_{mn}, \notag\\
	-\sum\limits_{m,n} (\kappa_x^2 m^2 + \kappa_y^2 n^2) C_{mn}\psi_{mn} & + k_0^2 \sum\limits_{l,p} \eps_{lp}\psi_{lp} \sum\limits_{m,n} C_{mn}\psi_{mn}\notag\\
	& + \sum\limits_{l,p} \hat{\eps}^{(x)}_{lp}\psi_{lp} \left( \sum\limits_{m,n} i \kappa_x m C_{mn}\psi_{mn} - \sum\limits_{m,n} i \kappa_y n B_{mn}\psi_{mn} \right) = k_z^2 \sum\limits_{m,n} C_{mn}\psi_{mn}.
\end{align}
Here
\begin{align}
	& {\eps}_{mn} = \frac{1}{L_x L_y} \int\limits_{-L_x/2}^{L_x/2} \int\limits_{-L_y/2}^{L_y/2} \frac{\eps(x,y)}{\psi_{mn}} dx dy, \label{eq10}\\
	& \hat{\eps}^{(x)}_{mn} = \frac{1}{L_x L_y} \int\limits_{-L_x/2}^{L_x/2} \int\limits_{-L_y/2}^{L_y/2} \dfrac{1}{\eps(x,y)}\dfrac{\partial\eps(x,y)}{\partial x} \frac{1}{\psi_{mn}} dx dy,\notag\\
	& \hat{\eps}^{(y)}_{mn} = \frac{1}{L_x L_y} \int\limits_{-L_x/2}^{L_x/2} \int\limits_{-L_y/2}^{L_y/2} \dfrac{1}{\eps(x,y)}\dfrac{\partial\eps(x,y)}{\partial y} \frac{1}{\psi_{mn}} dx dy.\notag
\end{align}

\begin{figure}[ht!]
	\centering
	\includegraphics[width=0.8\linewidth]{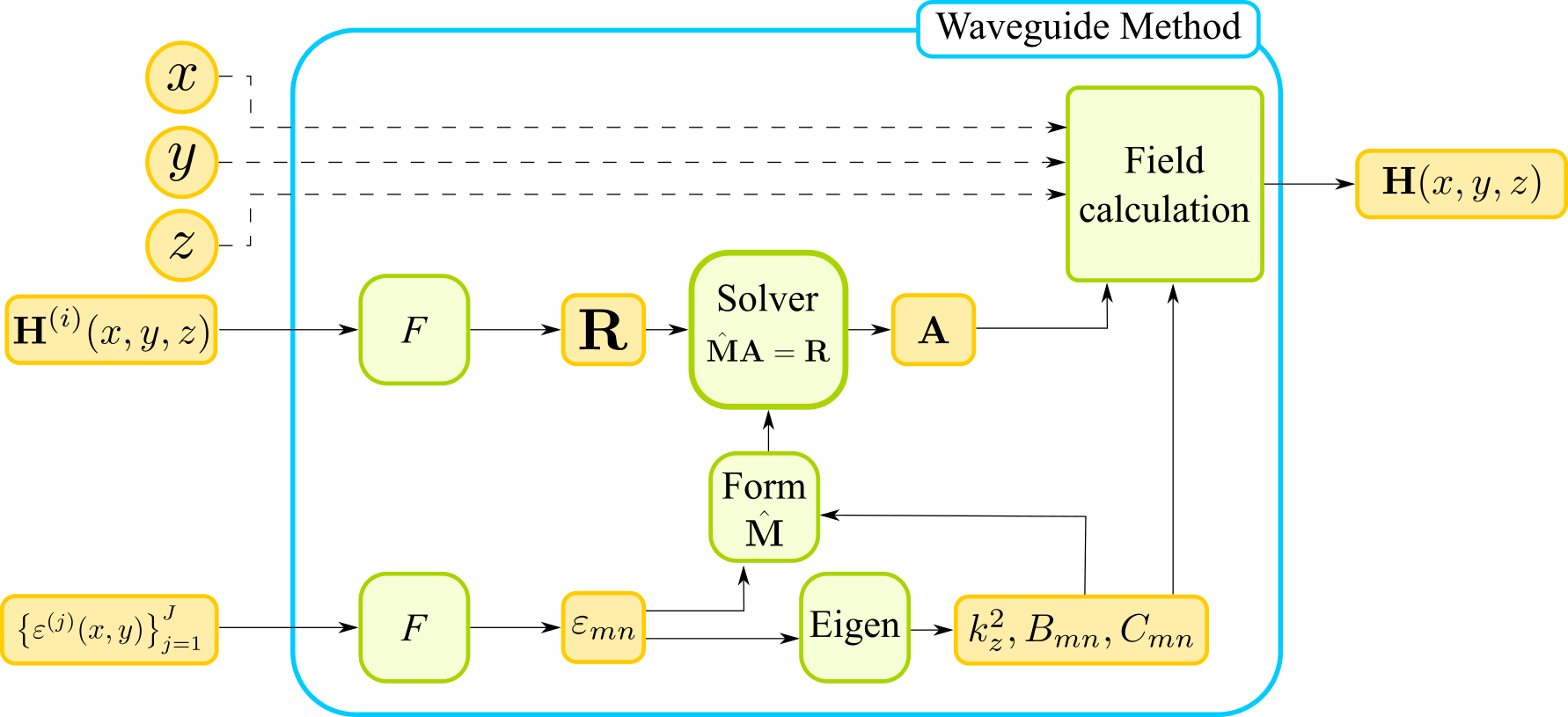}
	\caption{Schematic diagram of the waveguide method. Here ``F'' is forward Fourier transform, ``Eigen'' is calculating eigenvalues and eigenvectors of system (\ref{eq13}) for every layer, ``Field calculation'' is calculating field with Equations (\ref{eq14})~--~(\ref{eq16}).}
	\label{fig3}
\end{figure}

Simplifying the equations, we obtain:
\begin{align}
	-\sum\limits_{m,n} (\kappa_x^2 m^2 + \kappa_y^2 n^2) B_{mn}\psi_{mn} & + k_0^2 \sum\limits_{l,p}\sum\limits_{m,n} \eps_{lp}  B_{mn} \psi_{(m+l)(n+p)}\notag\\
	& - i \sum\limits_{l,p}\sum\limits_{m,n} \hat{\eps}^{(y)}_{lp} \left(  \kappa_x m C_{mn} - \kappa_y n B_{mn} \right)\psi_{(m+l)(n+p)} = k_z^2 \sum\limits_{m,n} B_{mn}\psi_{mn}, \notag\\
	-\sum\limits_{m,n} (\kappa_x^2 m^2 + \kappa_y^2 n^2) C_{mn}\psi_{mn} & + k_0^2 \sum\limits_{l,p}\sum\limits_{m,n} \eps_{lp}  C_{mn}\psi_{(m+l)(n+p)}\notag\\
	& + i \sum\limits_{l,p}\sum\limits_{m,n} \hat{\eps}^{(x)}_{lp} \left(  \kappa_x m C_{mn} - \kappa_y n B_{mn} \right)\psi_{(m+l)(n+p)} = k_z^2 \sum\limits_{m,n} C_{mn}\psi_{mn}.
\end{align}
Then, we obtain the following equations:
\begin{align}
	-(\kappa_x^2 m^2 + \kappa_y^2 n^2) B_{mn} & + k_0^2 \sum\limits_{l,p} \eps_{(m-l)(n-p)}  B_{lp}\notag\\
	& - i \sum\limits_{l,p} \hat{\eps}^{(y)}_{(m-l)(n-p)} \left(  \kappa_x l C_{lp} - \kappa_y p B_{lp} \right) = k_z^2 B_{mn}, \notag\\
	- (\kappa_x^2 m^2 + \kappa_y^2 n^2) C_{mn} & + k_0^2 \sum\limits_{l,p} \eps_{(m-l)(n-p)}  C_{lp}\notag\\
	& + i \sum\limits_{l,p}\hat{\eps}^{(x)}_{(m-l)(n-p)} \left(  \kappa_x l C_{lp} - \kappa_y p B_{lp} \right) = k_z^2 C_{mn}.\label{eq12}
\end{align}
In the system of equations (\ref{eq12}) the values $k_z^2$ must be read as eigenvalues, $B_{mn}$ and $C_{mn}$ are elements of eigenvectors of the matrix on the left side of these equations. The system of Equations (\ref{eq12}) is transformed into following large algebraic eigenvalue problem for each layer $j$ (see~\cite{lucas1996efficient})
\begin{align}
	\mathbf{M}^{(j)}_{\text{layer}} \begin{pmatrix} \mathbf{B}^{(j)} \\ \mathbf{C}^{(j)} \end{pmatrix} = \left(k_z^{(j)}\right)^2 \begin{pmatrix} \mathbf{B}^{(j)} \\ \mathbf{C}^{(j)} \end{pmatrix},\label{eq13}
\end{align}
where $\mathbf{M}^{(j)}_{\text{layer}}$ is the matrix derived from the left-hand side of Equations~(\ref{eq12}), $\mathbf{B}^{(j)}$ and $\mathbf{C}^{(j)}$ are the eigenvectors formed by $B_{mn}$ and $C_{mn}$. The eigenvalues $\left(k_z^{(j)}\right)^2$ give the propagation constants. The total field in each $j$th layer is a linear combination of modes
\begin{align}
	\begin{bmatrix} H_x^{(j)} \\ H_y^{(j)} \end{bmatrix} &{=} \sum\limits_{p = 1}^{2N} k_{z;p}^{(j)} \left[A_{p;1}^{(j)} e^{i k_{z;p}^{(j)} z} + A_{p;2}^{(j)} e^{-i k_{z;p}^{(j)} z}\right]\notag\\
	&\times \sum\limits_{m,n{=}-N_x,-N_y}^{N_x,N_y} \begin{bmatrix}{B}_{p,mn}^{(j)} \\ {C}_{p,mn}^{(j)}\end{bmatrix} \psi_{mn},\label{eq14}
\end{align}
where $N_x$ and $N_y$ are maximum numbers of the Fourier harmonics along the $x$ and $y$ axes, $N = (2 N_x + 1) (2 N_y + 1)$. The reflected field ($z>0$) is written as:
\vspace{-0.1cm}
\begin{align}
	\begin{bmatrix} H_x^{(r)} \\ H_y^{(r)} \end{bmatrix} {=} \hspace{-1mm}\sum\limits_{m,n = -N_x,-N_y}^{N_x,N_y} \hspace{-3mm}{k}_{z;mn} \begin{bmatrix} A_{x;mn}^{(r)} \\ A_{y;mn}^{(r)} \end{bmatrix}  \psi_{mn} e^{- i {k}_{z;mn} z}.\label{eq15}
\end{align}
The transmitted field ($z<-D$) is written as:
\vspace{-0.1cm}
\begin{align}
	\begin{bmatrix} H_x^{(t)} \\ H_y^{(t)} \end{bmatrix} {=} \hspace{-1mm}\sum\limits_{m,n = -N_x,-N_y}^{N_x,N_y} \hspace{-3mm}{k}_{z;mn} \begin{bmatrix} A_{x;mn}^{(t)} \\ A_{y;mn}^{(t)} \end{bmatrix}  \psi_{mn} e^{i {k}_{z;mn} z}.\label{eq16}
\end{align}
Here $A_{p;1}^{(j)}$, $A_{p;2}^{(j)}$, $A_{x;mn}^{(r)}$, $A_{y;mn}^{(r)}$, $A_{x;mn}^{(t)}$ and $A_{y;mn}^{(t)}$ are the unknown coefficients. By satisfying continuity conditions of the tangential field components at each layer interface (at $z=z_{\min;j}$ and $z=z_{\max;j}$ for $j$th layer)
\begin{eqnarray}
	&& \left.H_x^{(j-1)}\right|_{z=z_{\max;j}} = \left.H_x^{(j)}\right|_{z=z_{\max;j}},\quad \left.H_y^{(j-1)}\right|_{z=z_{\max;j}} = \left.H_y^{(j)}\right|_{z=z_{\max;j}},\notag\\ 
	&& \left.E_x^{(j-1)}\right|_{z=z_{\max;j}} = \left.E_x^{(j)}\right|_{z=z_{\max;j}},\quad \left.E_y^{(j-1)}\right|_{z=z_{\max;j}} = \left.E_y^{(j)}\right|_{z=z_{max;j}},\notag\\
	&& \left.H_x^{(j+1)}\right|_{z=z_{\min;j}} = \left.H_x^{(j)}\right|_{z=z_{\min;j}},\quad \left.H_y^{(j+1)}\right|_{z=z_{\min;j}} = \left.H_y^{(j)}\right|_{z=z_{\min;j}},\notag\\ 
	&& \left.E_x^{(j+1)}\right|_{z=z_{\min;j}} = \left.E_x^{(j)}\right|_{z=z_{\min;j}},\quad \left.E_y^{(j+1)}\right|_{z=z_{\min;j}} = \left.E_y^{(j)}\right|_{z=z_{\min;j}},\notag
\end{eqnarray}
a global system of linear equations is formed:
\begin{equation}\label{eq17}
	\hat{\mathbf{M}} \mathbf{A} = \mathbf{R},
\end{equation}
where $\hat{\bf M}$ is matrix of system of equations obtained from the boundary conditions, $\mathbf{A}$ is the vector of unknown mode amplitudes (noted above) and $\mathbf{R}$ is determined by the incident field. Solving this large linear system is the most computationally expensive part of the WG method. Figure~\ref{fig3} shows a visual summary of the standard Waveguide method. Here, the 'Solver' block is the primary computational bottleneck.

\subsection{Physics-Informed Neural Networks}
According to PINN approach~\cite{Raissi2019}, which relies on the universal approximation theorem~\cite{HORNIK1989359,Cybenko1989ApproximationBS}, we represent the latent variables $H_x$ and $H_y$ with a deep neural network ${\vec u}_{{\bm \theta}}({\vec x})$, where ${\bm \theta}$ denote all trainable parameters (e.g., {weights} and biases) of the network. We use multilayer perceptron (MLP) as deep neural network. Finding optimal parameters is an optimization problem, which requires {the definition} of a loss function such that its minimum gives the solution of the PDEs~(\ref{eq2}). The physics-informed model is trained by minimizing the composite loss function which consists of the local residuals of the differential equation over the problem domain and its boundary as shown below:
\begin{equation}\label{eq18}
	\mathcal{L}({\bm \theta}) = \lambda_{bc} \mathcal{L}_{bc}({\bm \theta}) + \lambda_{r} \mathcal{L}_{r}({\bm \theta}),	  
\end{equation}
where
\begin{eqnarray}
	&& \mathcal{L}_{bc}\left(\bm \theta \right) = \frac{1}{N_{bc}} \sum_{i=1}^{N_{bc}} \left| \mathcal{B}\left[{\vec u}_{{\bm \theta}} \right] \left({\vec x}_{i}^{(bc)}  \right)  \right|^{2}\label{eq19},\\
	&& \mathcal{L}_{r}\left(\bm \theta \right) = \frac{1}{N_r} \sum_{i=1}^{N_{r}} \left| \mathcal{R}\left[{\vec u}_{{\bm \theta}} \right] \left({\vec x}_{i}^{(r)} \right)\right|^{2},\label{eq20}\\
	&& \mathcal{R}\left[{u_1} \right]:= \Delta u_1 + k_0^2 \eps u_1 + \frac{1}{\eps} \frac{\partial \eps}{\partial y} \cdot \left( \frac{\partial u_2}{\partial x} - \frac{\partial u_1}{\partial y} \right),\\
	&& \mathcal{R}\left[{u_2} \right]:= \Delta u_2 + k_0^2 \eps u_2 +  \frac{1}{\eps}\frac{\partial \eps}{\partial x} \cdot \left( \frac{\partial u_1}{\partial y} - \frac{\partial u_2}{\partial x} \right).\label{eq21}
\end{eqnarray}
Here $\mathcal{L}_{bc}$ is a boundary loss term that corresponds to the boundary conditions, and $\mathcal{L}_{r}$ is a residual loss term that  corresponds to non-zero residuals of the governing PDEs (see details in~\cite{Raissi2019,Eskin2024}), $\lambda_{bc}$ and $\lambda_{r}$ are hyperparameters, which allow for separate tuning of the learning rate for each of the loss terms in order to improve the convergence of the model (see the procedure for choosing these parameters and examples~\cite{Raissi2019,Eskin2024,ESKIN2025114085}; in our calculations $\lambda_{bc} = N_x / L_x$ and $\lambda_{r}=1$), $\mathcal{B}$ is a boundary operator corresponding to boundary conditions:
\begin{align}
	& {\vec u}(x,y,z) = {\vec u}(x+L_x,y,z), \notag\\
	& {\vec u}(x,y,z) = {\vec u}(x,y+L_y,z), \notag\\
	& \left.\int\limits_{-L_x/2}^{L_x/2} \int\limits_{-L_y/2}^{L_y/2} \left(\frac{\partial {\vec u}}{\partial z} + i {k}_{z;mn} {\vec u} \right) \frac{1}{\psi_{mn}} dx dy\right|_{z = z_{\max}} = 0, \notag\\
	& \left.\int\limits_{-L_x/2}^{L_x/2} \int\limits_{-L_y/2}^{L_y/2} \left(\frac{\partial {\vec u}}{\partial z} - i {k}_{z;mn} {\vec u} \right) \frac{1}{\psi_{mn}} dx dy\right|_{z = z_{\min}} = 0, \label{eq23}
\end{align}
where the first two conditions are the periodicity condition of the solution, the last two are the Sommerfeld conditions at the upper ($z_{\max}$) and lower ($z_{\min}$) boundaries of the domain under consideration. 
$\left\{{\vec x}_{i}^{(bc)}\right\}^{N_{bc}}_{i=1}$ and $\left\{{\vec x}_{i}^{(r)}\right\}^{N_{r}}_{i=1}$ are sets of points corresponding to boundary condition domain and PDE domain, respectively.
These points can be the vertices of a fixed mesh or can be randomly sampled at each iteration of a gradient descent algorithm. All required gradients w.r.t. input variables ($\vec x$) and network parameters (${\bm \theta}$) can be efficiently computed via automatic differentiation~\cite{Griewank2008} with algorithmic accuracy, which is defined by the accuracy of computation system. 

\begin{figure}[ht!]
	\centering
	\includegraphics[width=0.8\linewidth]{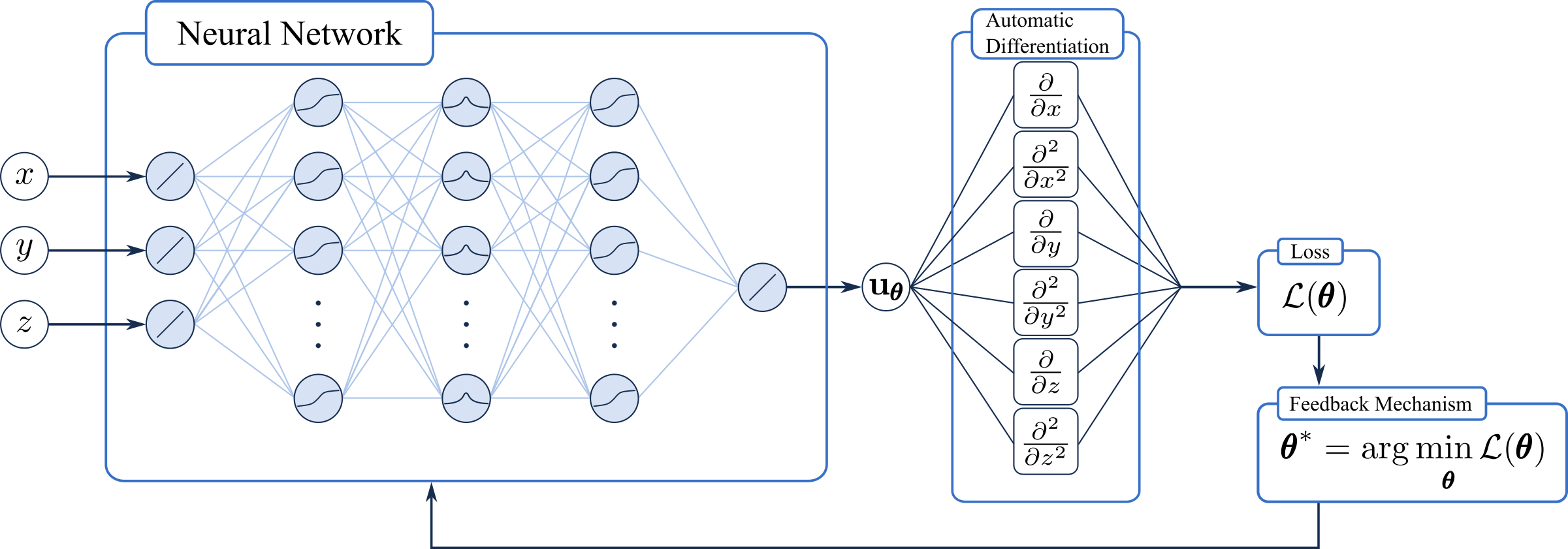}
	\caption{Schematic diagram of the general PINN method.}
	\label{fig4}
\end{figure}

The optimization problem can be defined as follows:
\begin{equation}
	{\bm \theta}^* = {\arg}\,\underset{{\bm \theta}}{\min} \, \mathcal{L}({\bm \theta}),\label{eq9}
\end{equation}
where ${\bm \theta}^*$ are optimal parameters of the neural network which minimize the discrepancy between the exact unknown solution ${\vec u}$ and the approximate one ${\vec u}_{\pmb \theta^*}$. A schematic diagram of the general PINN method, which is used throughout this paper, is shown Figure~\ref{fig4}. This neural network learning method is unsupervised, meaning that training is carried out without labeled datasets.

\subsection{Waveguide Neural Operator}
The development and use of neural networks primarily focus on training mappings between finite-dimensional Euclidean spaces or finite sets. This approach has been extended to neural operators that learn mappings between functional spaces~
\cite{lu2021learning,li2020neuraloperatorgraphkernel,li2021fourierneuraloperatorparametric,yang2022genericlithographymodelingdualband}. For partial differential equations, neural operators directly learn a mapping from any given functional parametric dependence to a solution, thereby learning an entire class of PDE solutions, as opposed to classical methods which solve one specific instance of an equation.

The process of solving the given problem can be considered as operator $\mathcal{G}: \mathcal{A} \rightarrow \mathcal{U}$ between function spaces $\mathcal{A}$ and $\mathcal{U}$. In the case of the mask problem, $\mathcal{A}$ and $\mathcal{U}$ are an infinite function space of functions of incident field and an infinite function space of functions of reflected, transmitted and in-mask fields, respectively.

The WG method is solution operator $\mathcal{G}^{\dagger}$ of the given diffraction problem:
\begin{align}
	\mathcal{G}^{\dagger}: \mathcal{A} \rightarrow \mathcal{U}.
\end{align} 
We aim to approximate $\mathcal{G}^\dagger$ by constructing a parametric map
\begin{align}
	\mathcal{G}_{\pmb \theta}: \mathcal{A} \rightarrow \mathcal{U}, \quad {{\pmb \theta}} \in \Theta,
\end{align} 
for some finite-dimensional parameter space $\Theta$ by choosing ${{\pmb \theta}}^{\dagger} \in \Theta$ so that $\mathcal{G}_{{\pmb \theta}^{\dagger}} \approx \mathcal{G}^{\dagger}$.

\begin{figure}[ht!]
	\centering
	\includegraphics[width=0.8\linewidth]{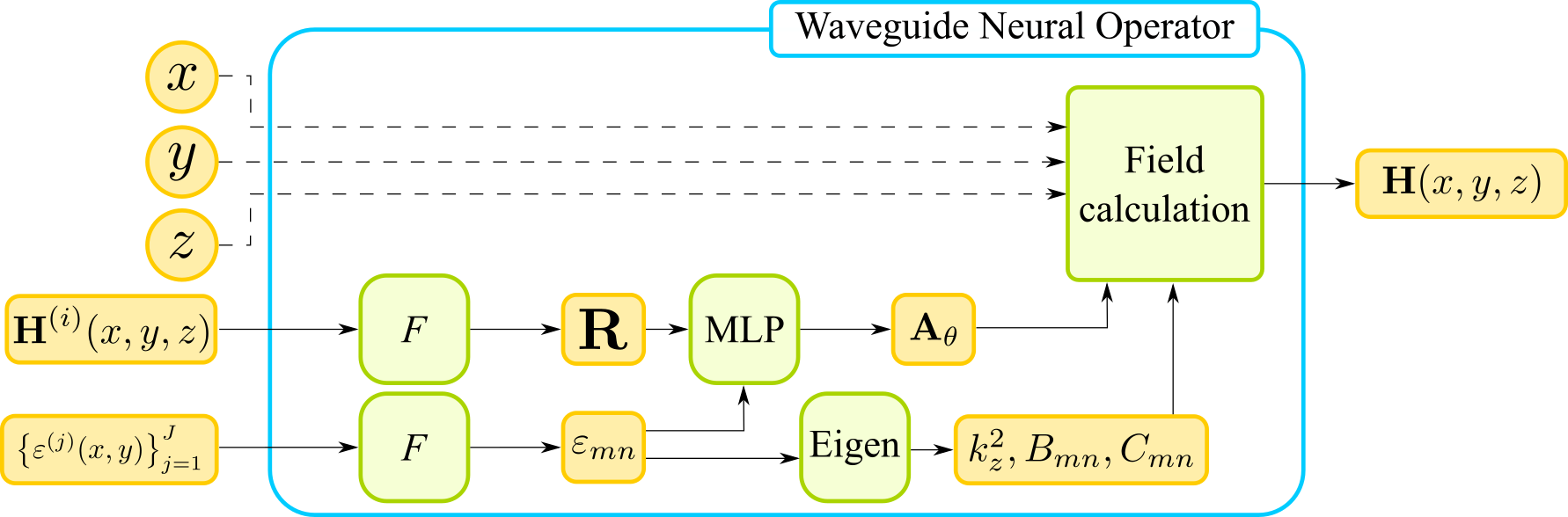}
	\caption{Schematic diagram of the waveguide neural operator. Here ``F'' is forward Fourier transform, ``Eigen'' is calculating eigenvalues and eigenvectors of system (\ref{eq13}) for every layer, ``Field calculation'' is calculating field with Equations (\ref{eq14})~--~(\ref{eq16}).}
	\label{fig5}
\end{figure}

Let's construct $\mathcal{G}_{\pmb \theta}$ from $\mathcal{G}^{\dagger}$ by replacing the most resource-intensive part of $\mathcal{G}^{\dagger}$ with a deep neural network. Specifically, we train a multi-layer perceptron (MLP) to learn the mapping from the right-hand side vector $\mathbf{R}$ and the mask parameters (Fourier coefficients of permittivity) to the solution vector $\mathbf{A}$ of the linear system (\ref{eq17}):
\begin{equation}
	\mathbf{A}_{\bm{\theta}} = \text{MLP}\left(\mathbf{R}, \{\varepsilon_{mn}^{(j)}\}^{J}_{j=1}; \bm{\theta}\right),
\end{equation}
where ${\eps}^{(j)}_{mn}$  is given by Equation~(\ref{eq10}). The optimization problem can be defined as follow:
\begin{align}
	& {\bm \theta}^{\dagger} = \arg\underset{{\bm \theta}}{\min} \mathcal{L}({\bm \theta}),\label{eq29_}\\
	&\mathcal{L}({\bm \theta}) = \|\hat{\mathbf{M}} \mathbf{A}_{\bm{\theta}} - \mathbf{R}\|^2_2. \notag
\end{align}
In this formulation of the problem, the neural operator $\mathcal{G}_{\pmb \theta}$ is mesh--independent, and its training takes place in a latent (Fourier) space, inherits the physical structure of the WG method, and directly targets the primary computational bottleneck. A schematic diagram of the WGNO is shown Figure~\ref{fig5}.

\section{Numerical Experiments}

We evaluated the performance of the proposed methods on several test problems. {The ${\rm TE}$ polarization of the incident waves was used for all considered cases.} The neural networks were implemented in PyTorch~\cite{Paszke2019} (version 2.6 under CUDA 12.4) and the training was carried out on a node with GPU Nvidia GeForce GTX 1660 Ti and CPU Intel Core i5-9300H. For {the} PINNs, we used a MLP with 2 hidden layers and 128 neurons each with hyperbolic tangent activation function ($\tanh$). To save computational resources and to accelerate convergence for problems with periodic boundary conditions along $x$ (in 2D case) it was constructed a Fourier feature embedding of input coordinates in the following form
\begin{equation}\label{eq29}
	v(x) = \left[1,\cos(k_x x),\sin(k_x x),\cos(2 k_x x),\sin(2 k_x x),...,\cos(m k_x x),\sin(m k_x x)\right],
\end{equation}
where $k_x = 2\pi/L_x$ and $m$ is some non-negative integer. With such embedding any trained approximation of problem solution by PINN exactly satisfies periodic conditions (see details in~\cite{Eskin2024,WANG2024116813,DONG2021110242}). For WGNO, the MLP had 2 hidden layers.

\subsection{Validation on Test Problems}
To evaluate the accuracy of the solutions provided by numerical solvers (FreeFem++, WG) and solvers based on artificial neural networks, we have tested the methods on three 2D problems with known analytical solutions: (1) plane wave propagation in a homogeneous medium, (2) reflection from a single interface (Figure~\ref{fig6} (a)), and (3) reflection from a dielectric layer (Figure~\ref{fig6} (b)). As the main performance metrics for the considered solvers, we evaluated the relative ${\mathbb L}_2$ error (see~\cite{Eskin2024} and appendix~{\ref{App1}) and the time to provide a solution (the time to calculate a solution for numerical solver or the inference time for neural networks). In addition, training times are marked separately for neural networks.
	
\begin{figure}[ht!]
	\centering
	\includegraphics[width=0.4\linewidth]{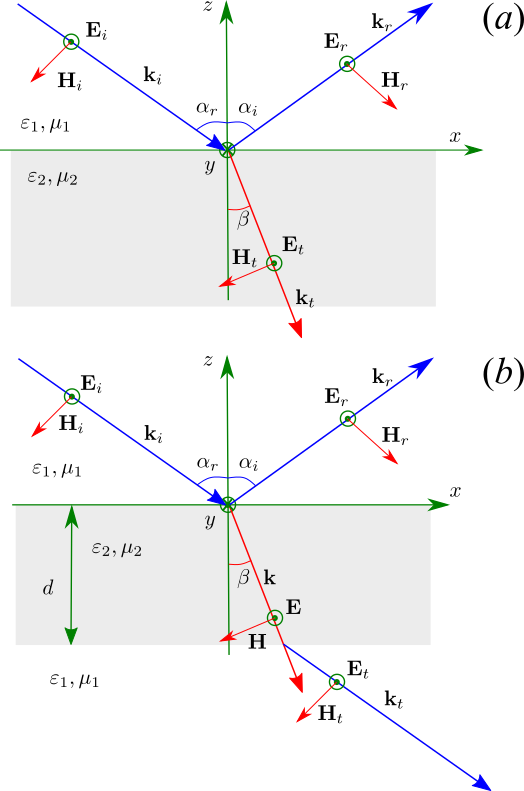}
	\caption{Problems with exact solutions: (a) is problem of electromagnetic waves reflection from a single interface, (b) is problem of electromagnetic waves reflection from a dielectric layer.}
	\label{fig6}
\end{figure}

We assume the media are homogeneous in the $y$ direction. An electromagnetic wave of TE polarization propagates in free space with wave vector ${\vec k}= k_x^{(i)} {\vec x}_0 - k_z^{(i)} {\vec z}_0$ ($k_x^{(i)}$ and $k_z^{(i)}$ are positive values, and $k_x^{(i)} = k_z^{(i)}$). The medium below interface and in the dielectric layer have permittivity $\varepsilon = 4$. The dielectric layer thickness is $\pi/ k_z^{(i)}$. The coordinates $(x, z)$ of the points in the computational domain satisfy the relations: $x \in [-\pi/ k_x^{(i)}, \pi/ k_x^{(i)}]$ and $z \in [-\pi/ k_z^{(i)}, \pi/ k_z^{(i)}]$ ($z \in [-2\pi/ k_z^{(i)}, \pi/ k_z^{(i)}]$ for the dielectric layer).

For the numerical solver we divided every axis into $100$ intervals, as a result we had about $10^4$ collocation points. $N_x$ equals $10$ for WG and WGNO methods at all test experiments.

For the PINN and {the} WGNO approach we divided the segment along every axis into $100$ intervals. In our experiments for the PINN, we used two-stage learning which consisted of 1000 epochs of optimization with the Adam optimizer with learning rate $10^{-3}$ and 5 epochs of optimization with the LBFGS optimizer. For WGNO we used 1000 epochs of optimization with the Adam optimizer with learning rate $10^{-3}$ and  $10^{-5}$. We trained randomly initialized neural networks using the Glorot scheme~\cite{Glorot10a}, repeating the training 7 times with different random seeds.

The results are summarized in Table~\ref{tab:test_problems}. The WG method achieves machine precision, as expected. {FreeFem++ provides sufficient computational accuracy for applied (engineering) needs.} The PINN achieves reasonable accuracy, {but it is less accurate than FEM in the case of the third problem (dielectric layer)}. The proposed WGNO demonstrates high performance, with errors several orders of magnitude smaller than the PINN, highlighting the effectiveness of the hybrid physics-based approach. Comparisons of the electric field distributions of the exact solution and the solution given by solvers are given in Appendix~\ref{App2}. The fields are normalized to the electric field amplitude of the incident wave.

\begin{table*}[h!]
	\centering
	\caption{Performance comparison on 2D test problems. Inference times are $1.2\times 10^{-3}$ s and $1.7\times 10^{-4}$ s for {the} PINN and {the} WGNO, respectively.}
	\label{tab:test_problems}
	\begin{tabular}{l|cc|cc}
		\toprule
		\multirow{2}{*}{Problem} & \multicolumn{2}{c|}{Waveguide} & \multicolumn{2}{c}{FreeFem++} \\
		& Rel. $L_2$ Error & Time (s) & Rel. $L_2$ Error & Time (s) \\
		\midrule
		1. Wave propagation & $2.814\times 10^{-15}$ & $1.0\times 10^{-3}$ & $2.323\times 10^{-3}$ & 3.467 \\
		2. Wave scattering on interface & $2.730\times 10^{-15}$ & $2.0\times 10^{-3}$ & $3.818\times 10^{-3}$ & 3.385 \\
		3. Reflection from layer & $6.110\times 10^{-15}$ & $6.0\times 10^{-3}$ & $8.085\times 10^{-3}$ & 5.033 \\
		\midrule \midrule
		\multirow{2}{*}{Problem} & \multicolumn{2}{c|}{PINN} & \multicolumn{2}{c}{WGNO} \\
		& Rel. $L_2$ Error & Training (s) & Rel. $L_2$ Error & Training (s) \\
		\midrule
		1. Wave propagation & $1.694{\times} 10^{-4}  {\pm} 5.3 {\times} 10^{-5}$ & $855$ & $4.032{\times} 10^{-8}{\pm}
		1.6{\times} 10^{-8}$ & $3.7$ \\
		2. Wave scattering on interface & $1.596{\times} 10^{-3} {\pm} 7.9 {\times} 10^{-4}$ & $1532$ & $1.785{\times} 10^{-7} {\pm} 2.8 {\times} 10^{-7}$ & $3.7$ \\
		3. Reflection from layer & $5.507{\times} 10^{-2} {\pm} 4.4{\times} 10^{-3}$ & $1221$ & $4.709{\times} 10^{-5}{\pm}
		5.3 {\times} 10^{-5}$ & $11$ \\
		\bottomrule
	\end{tabular}
\end{table*}

\subsection{2D Lithography Mask Simulation}
We simulated a realistic 2D EUV mask (see Fig.~\ref{fig2}) for wavelengths ($\lambda$) of 13.5 and 11.2 nm.  These wavelengths correspond to the operating wavelengths of current industrial lithography systems~\cite{yakshin2000determination} and of promising ones~\cite{Chkhalo2024,Abramov2025}.

\subsubsection{Wavelength 13.5 nm}\label{section2DWDNO}
The mask included an absorber consisting of a TaBO layer and a TaBN layer  (10 nm for PINN, and 60 nm for WGNO), a thin Ru layer,  a Mo/Si layer for PINN and a 31-layer Mo/Si mirror for the WGNO (see Figure~\ref{fig2} (b)). The parameters of these masks are given in Table~\ref{table2}. The absorbers had a hole with a width $L_x/2$ so the permittivities of the absorber in layer $j$ ($j$ is 1 or 2) are described by:
\begin{align}
	\eps_j(x) &= \frac{1}{2}\left[\tanh\left(\frac{x + a}{d}\right) - \tanh\left(\frac{x - a}{d}\right)\right] (1 - \eps) + \eps. \label{eq30}
\end{align}
Here $\eps$ is dielectric permittivity of layer, $2 a$ is length of hole ($a=L_x / 4$) and $d$ is the thickness of the transition region at the edge of the hole ($d = \lambda / 10$). Note that we obtained the permittivities of the media from experimental data~\cite{HENKE1993181,CenterXRayOpt}.The angle of incidence of the wave is $6^{\circ}$, which corresponds to the angle of incidence of the waves on the mask in modern lithographic systems. As can be seen from Figure~\ref{fig7}, which shows the dependence of the reflection coefficient (power of reflected waves normalized to the power of incidence waves) on the angle of incidence of TE EUV wave for Mo/Si mirror of 60 layers, this angle lies on the plateau of the maximum of this dependence.

\begin{figure}[ht!]
	\centering
	\includegraphics[width=0.8\linewidth]{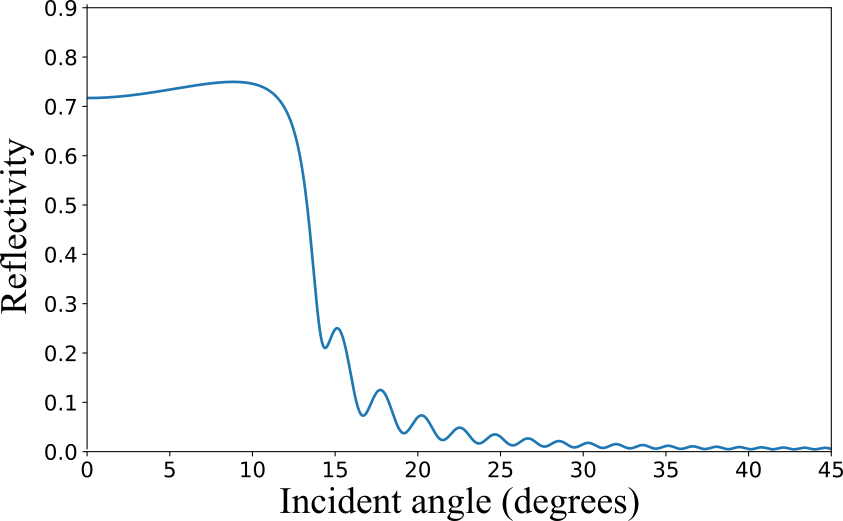}
	\caption{Dependence of reflectivity of TE EUV wave (wavelength 13.5 nm) on incident angle for Mo/Si mirror of 60 layers. Thicknesses of Mo and Si layers are given in Table~\ref{table2}. Maximum of reflectivity is 0.75 at incident angel 8.86\degree.}
	\label{fig7}
\end{figure}

Figures~\ref{fig_2d_PINN} and \ref{fig_2d_WGNO} show the comparisons of the calculated $E_y$ field, normalized to the amplitude of the incident wave, predicted by PINN and WGNO, respectively. It follows from the demonstrated results that while the PINN captures the general wave pattern, its absolute error is significant. In contrast, the WGNO result is visually indistinguishable from the reference WG solution, with a very small absolute error (see Fig.~\ref{fig_2d_WGNO}).

\begin{table}
	\centering
	\caption{Parameters of the mask for wavelength $\lambda{=}13.5$ nm which used to calculations.}
	\label{table2}
	\begin{tabular}{ll|ccccc}
		\toprule
		Method & Properties & \multicolumn{5}{c}{Media} \\
		       & Layer  & TaBO  & TaBN & Ru & Mo & Si \\
		\midrule
		 & $\varepsilon$ & $0.857{-}i 0.079$ & $0.861{-}i 0.071$ & $0.785{-}i 0.030$ & $0.853{-}i0.012$ & $0.998{-}i 0.004$\\
		PINN & thickness & 10 nm & 10 nm  & 2  nm & 3 nm & 4 nm \\
		WGNO & thickness & 10 nm & 60 nm  & 2  nm & 3 nm & 4 nm \\
		\bottomrule
	\end{tabular}
\end{table}

\begin{figure}[t!]\centering
	\includegraphics[width=0.8\textwidth]{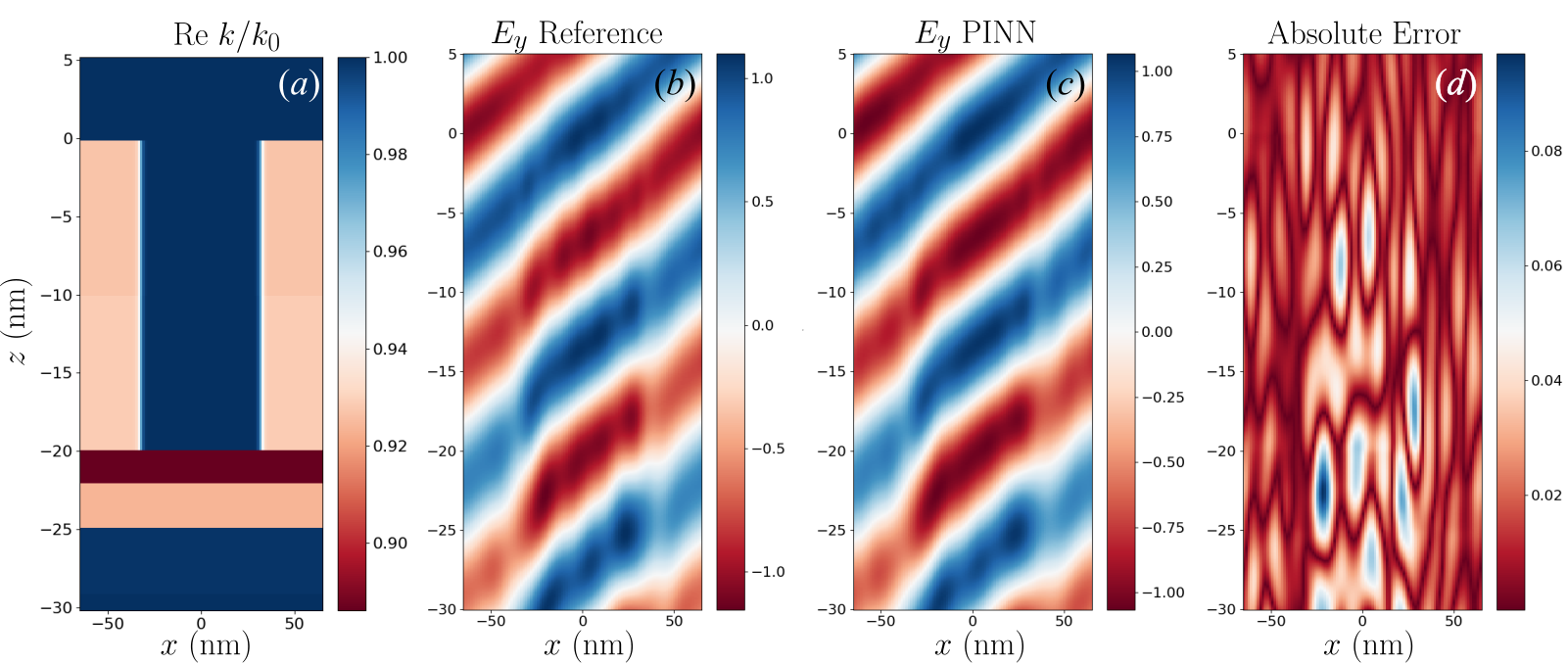}
	\caption{(a) is real part of $k$ in the media normalized to $k_0$. Comparison of reference WG solution (b) and PINN solution (c) for a 2D mask at 13.5 nm, (d) is absolute error.}\label{fig_2d_PINN}
\end{figure}

\begin{figure}[t!]\centering
	\includegraphics[width=0.8\textwidth]{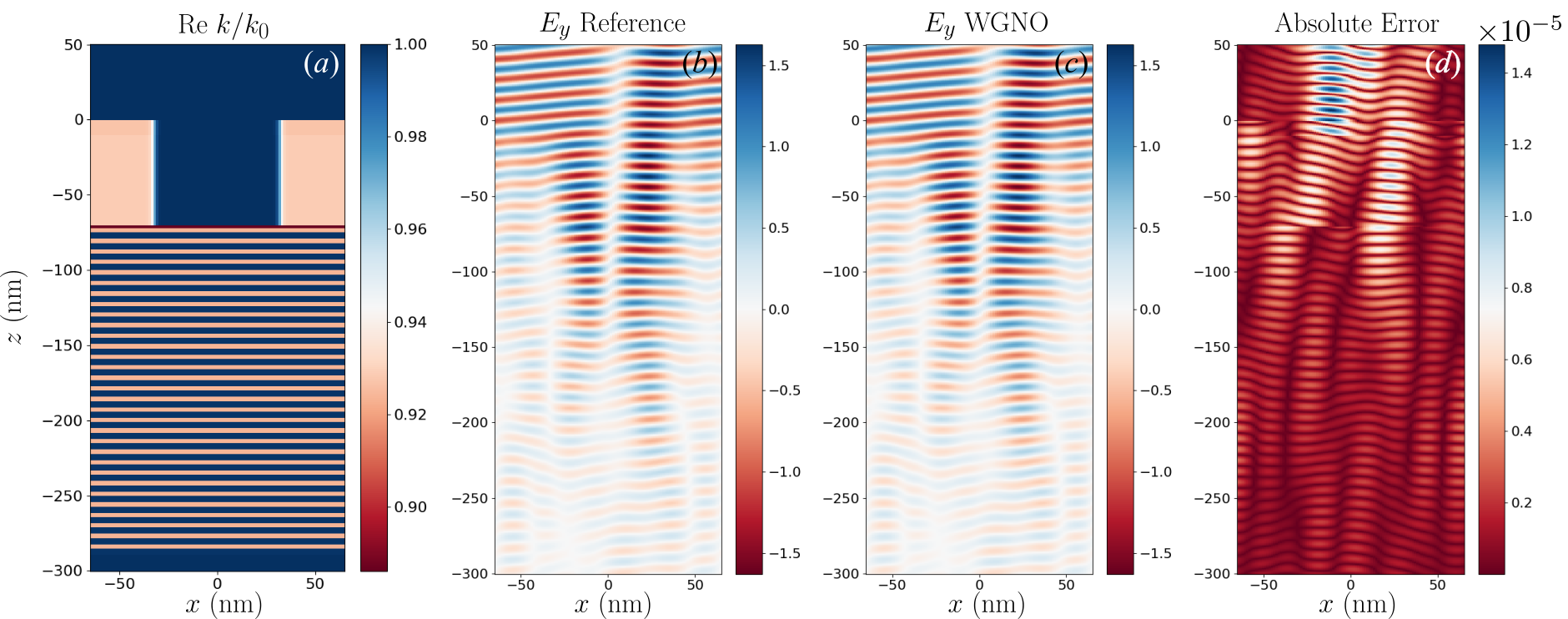}
	\caption{(a) is real part of $k$ in the media normalized to $k_0$. Comparison of reference WG solution (b) and WGNO solution (c) for a 2D mask at 13.5 nm, (d) is absolute error.}\label{fig_2d_WGNO}
\end{figure}

\subsubsection{Wavelength 11.2 nm}

For the wavelength of 11.2~nm, the top two layers of the mask follow the same configuration as described above: an absorber consisting of a TaBO layer (10~nm) and a TaBN layer (10~nm for PINN, 60~nm for WGNO). Since a standardized solution for mirrors at this wavelength has not yet been established and related research is still ongoing~\cite{Chkhalo17,Polkovnikov_2020,Shaposhnikov22}, we considered several potential mirror designs for further experiments. Figure~\ref{fig_R_on_theta_11_2} shows the reflectivity as a function of the angle of incidence for Mo/Be, Ru/Be, and Ru/Be/Sr mirrors, each consisting of 60 layers with thicknesses specified in Table~\ref{table3}. As can be seen, the Ru/Be/Sr mirror exhibits the highest reflectivity. Consequently, this structure was chosen as the mask substrate (see Figure~\ref{fig2} (c)). For the PINN-based solver, a single layer of this structure was utilized, whereas for the WGNO, the mirror consisted of a 31-layer Ru/Be/Sr. The dielectric permittivities of TaBO and TaBN are $\varepsilon=0.906 {-} i 0.064$ and $\varepsilon=0.909 {-} i 0.060$ respectively. As in the previous mask case, the absorbers had a hole with width $L_x/2$ so the permittivities of the absorbers are described by Eqn.~(\ref{eq29}). The angle of incidence of the wave is~$6^{\circ}$. Figures~\ref{fig_2d_PINN_11_2} and \ref{fig_2d_WGNO_11_2} show the comparisons of the calculated $E_y$ field, normalized to the amplitude of the incident wave, predicted by PINN and WGNO, respectively, for the wavelength 11.2 nm.

\begin{table}
	\centering
	\caption{Parameters of layers of mirror for wavelength $\lambda{=}11.2$ nm.}
	\label{table3}
	\begin{tabular}{ll|cccc}
		\toprule
		Mirror & Properties & \multicolumn{4}{c}{Media} \\
			   & Layer  & Mo & Ru & Be & Sr \\
		\midrule
		& $\varepsilon$ & $0.91{-}i 0.009$ & $0.872{-}i 0.012$ & $1.025{-} i 0.003$ & $0.986{-} i 0.002$\\
		Mo/Be    & thickness & 2.22 nm  & -- & 3.5 nm & -- \\
		Ru/Be    & thickness & --  & 2.01  nm & 3.72 nm & -- \\
		Ru/Be/Sr & thickness & --  & 1.7  nm & 2.7 nm & 1.34 nm \\
		\bottomrule
	\end{tabular}
\end{table}

\begin{figure}[ht!]
	\centering
	\includegraphics[width=0.8\linewidth]{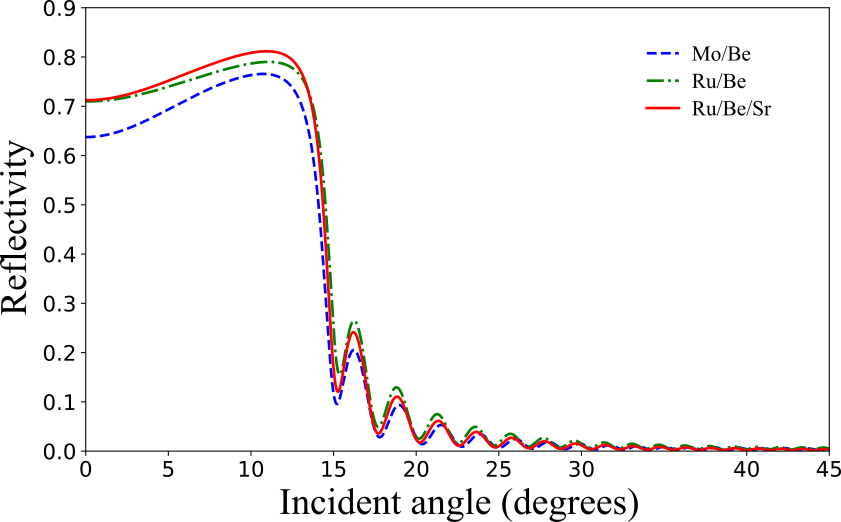}
	\caption{Dependences of reflectivity of TE EUV wave (wavelength 11.2 nm) on incident angle for Mo/Be, Ru/Be and Ru/Be/Sr mirrors of 60 layers. Thicknesses of layers are given in Table~\ref{table3}. Maximum of reflectivities are: 0.77 at incident angel 10.71\degree for Mo/Be, 0.79 at incident angel 11.06\degree for Ru/Be, and 0.81 at incident angel 10.96\degree for Ru/Be/Sr.}
	\label{fig_R_on_theta_11_2}
\end{figure}

\begin{figure}[t!]\centering
	\includegraphics[width=0.8\textwidth]{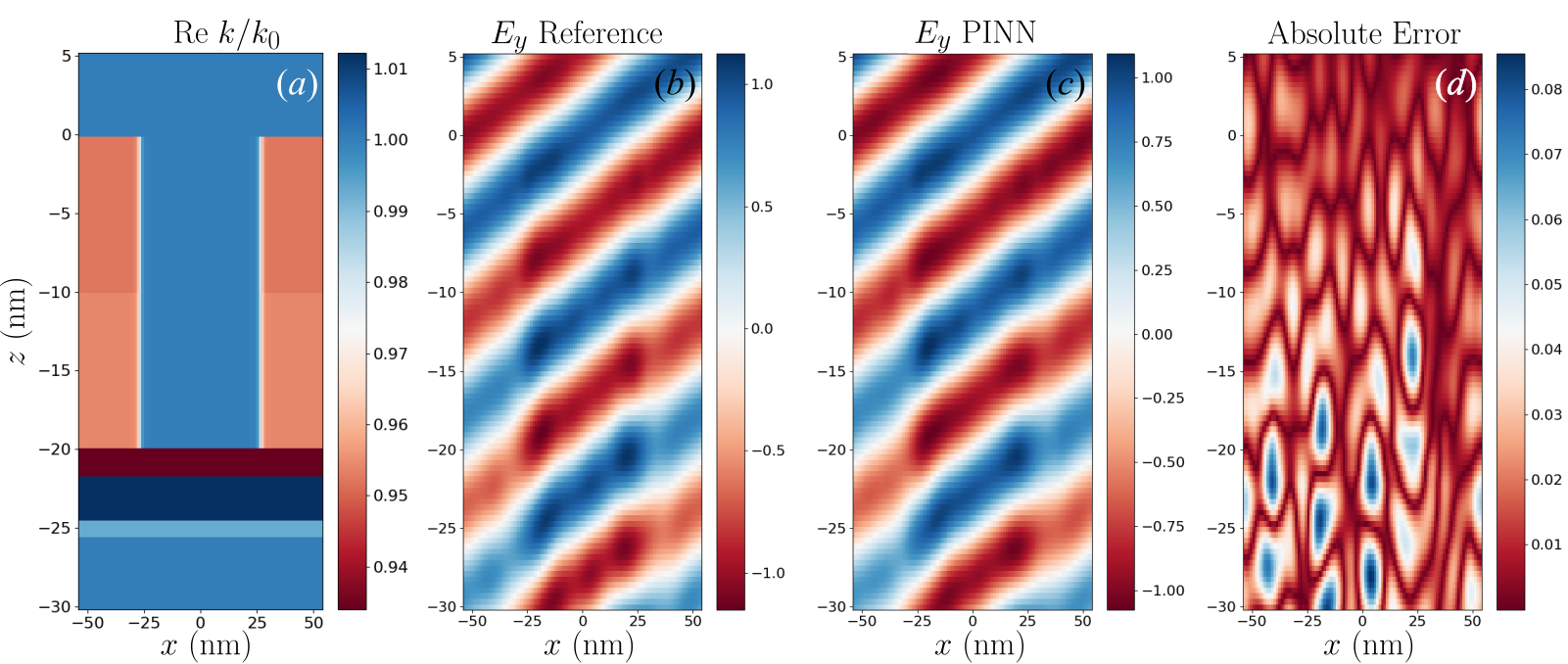}
	\caption{(a) is real part of $k$ in the media normalized to $k_0$. Comparison of reference WG solution (b) and PINN solution (c) for a 2D mask at 11.2 nm, (d) is absolute error.}\label{fig_2d_PINN_11_2}
\end{figure}

\begin{figure}[t!]\centering
	\includegraphics[width=0.8\textwidth]{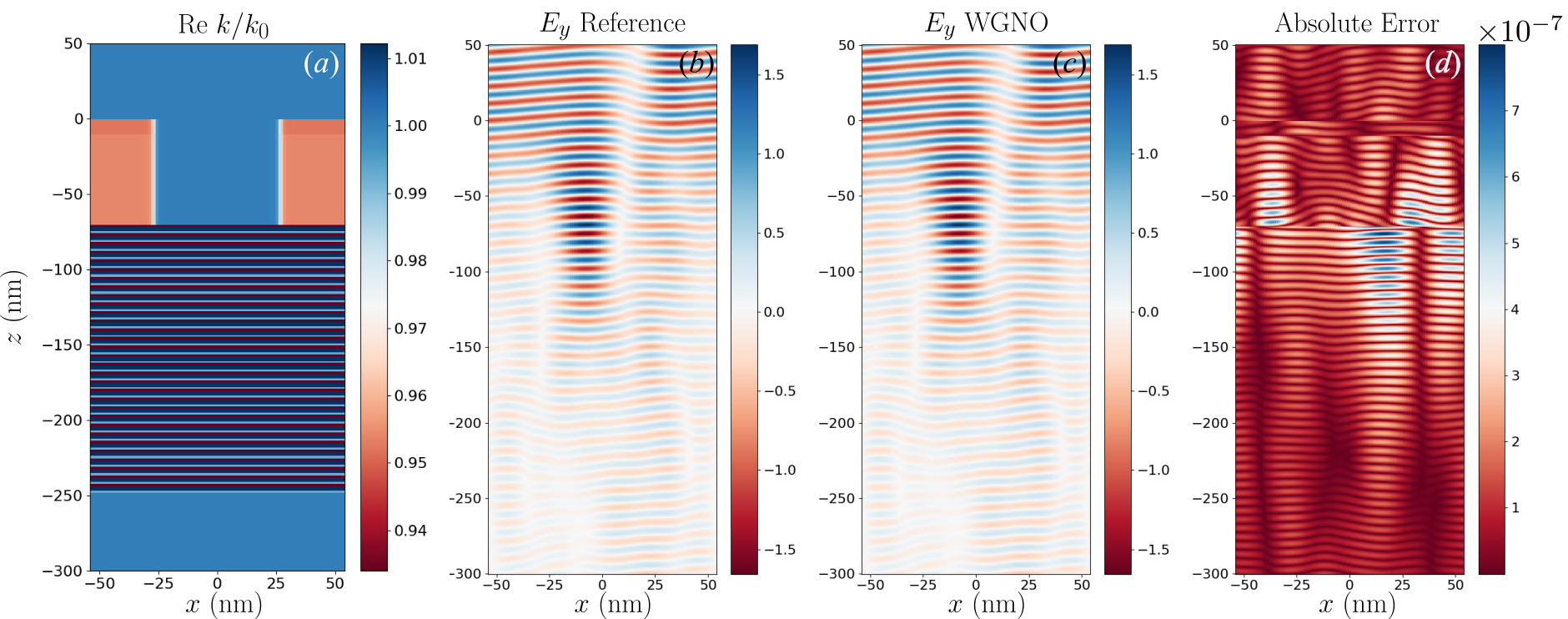}
	\caption{(a) is real part of $k$ in the media normalized to $k_0$. Comparison of reference WG solution (b) and WGNO solution (c) for a 2D mask at 11.2 nm, (d) is absolute error.}\label{fig_2d_WGNO_11_2}
\end{figure}

The quantitative results for the both wavelengths are summarized in Table~\ref{tab:2d_mask}. The PINN struggles with this complex problem, yielding errors of several percent after long training times. The WGNO, however, achieves excellent accuracy with a short training time.

\begin{table}[h!]
	\centering
	\caption{Performance on 2D lithography mask simulation. Inference times are $1.2\times 10^{-3}$ s and $1.8\times 10^{-4}$ s for PINN and WGNO, respectively.}
	\label{tab:2d_mask}
	\begin{tabular}{l|ccc}
		\toprule
		$\lambda$ (nm) & Method & \!Rel. \!\!$L_2$\! Error & \! Training (s)\\
		\midrule
		\multirow{2}{*}{13.5} & PINN & $3.1{\times} 10^{-2}$ & $1003$ \\
		& WGNO & $3.8{\times} 10^{-6}$ & $4.3$ \\
		\midrule
		\multirow{2}{*}{11.2} & PINN & $3.3{\times} 10^{-2}$ & $1252$ \\
		& WGNO & $3.5{\times} 10^{-7}$ & $7.7$  \\
		\bottomrule
	\end{tabular}
\end{table}

\subsection{3D Lithography Mask Simulation}
Finally, we extended our analysis to the full, challenging 3D mask problem. The mask structure and parameters of the incident field are as in previous case for the WGNO. The nonuniform permittivities of the absorber in layer $j$ are described by:
\begin{align}
	\eps_j(x,y) & = \frac{1}{4}\left[\tanh\left(\frac{x {+} a}{d}\right) {-} \tanh\left(\frac{x {-} a}{d}\right)\right]\notag\\
	& \times \left[\tanh\left(\frac{y {+} b}{d}\right) {-} \tanh\left(\frac{y {-} b}{d}\right)\right] (1 {-} \eps) {+} \eps,
\end{align}
where $\eps$ is permittivity of absorber of $j$th layer, $a=L_x / 4$, $b=L_y / 4$, $d = \lambda / 10$.

Figures~\ref{fig3D_13_5nm} and~\ref{fig3D_11_2nm} show cross-sections of the field calculated by our WGNO compared to the reference solution at 13.5 nm and 11.2 nm, respectively. The agreement is excellent, demonstrating the capability of our method to handle the full 3D problem with high fidelity.

The performance metrics for the 3D case, shown in Table~\ref{tab:3d_mask}, are even more compelling. The WGNO maintains high accuracy while achieving a speedup of over 200 times compared to the rigorous WG solver. The entire training process for the 3D case took less than 20 seconds, demonstrating the remarkable efficiency and scalability of our proposed operator.

\begin{figure}[t!]\centering
	\includegraphics[width=0.5\textwidth]{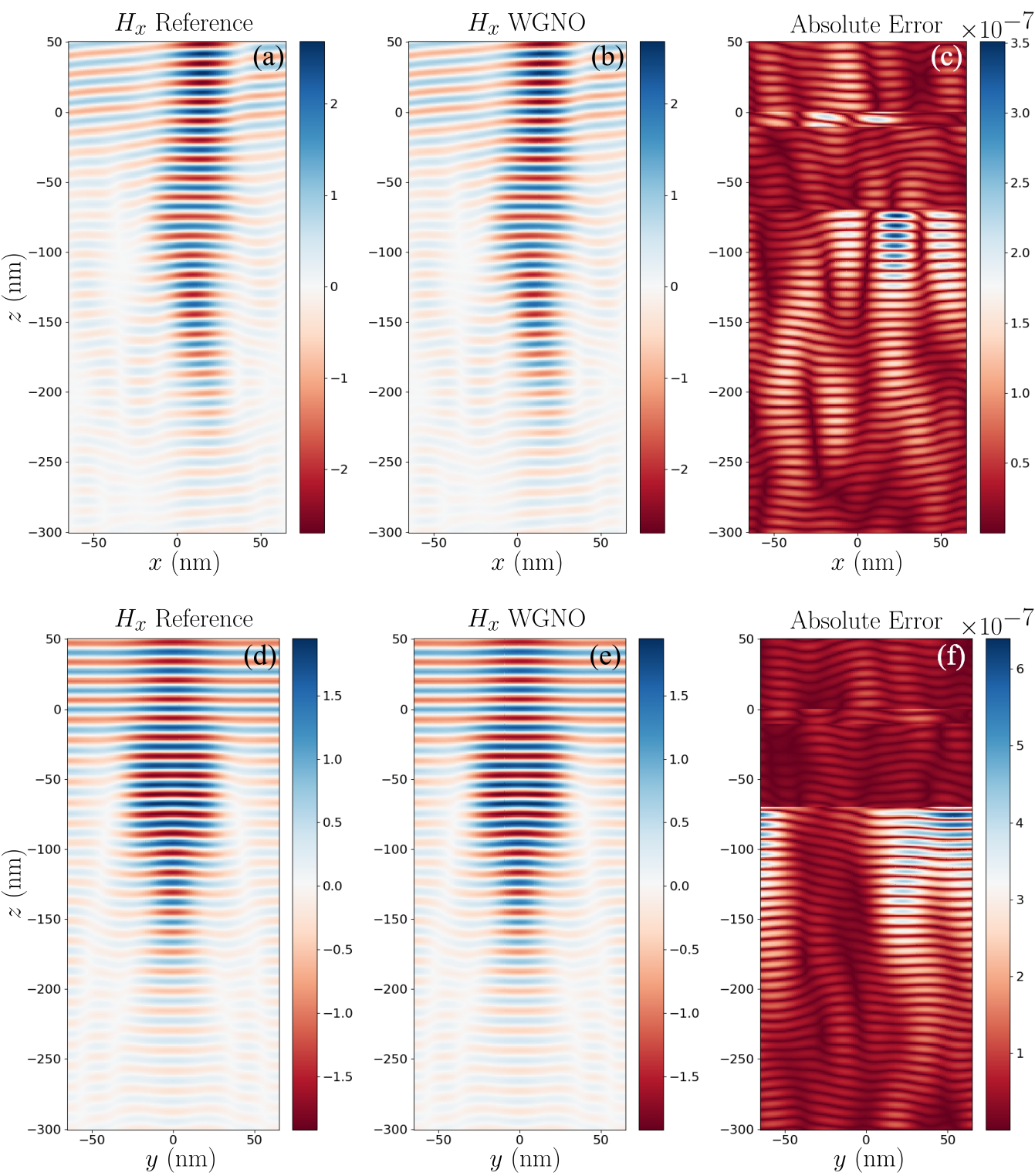}
	\caption{Comparison of reference WG solution ((a),(d)) and WGNO solution ((b),(e)) for a 3D mask at 13.5 nm, (c) and (f) are absolute errors. (a)--(c) are in cross-section $y=0$, (d)--(f) are in the cross-section $x=0$.}\label{fig3D_13_5nm}
\end{figure}

\begin{figure}[t!]\centering
	\includegraphics[width=0.5\textwidth]{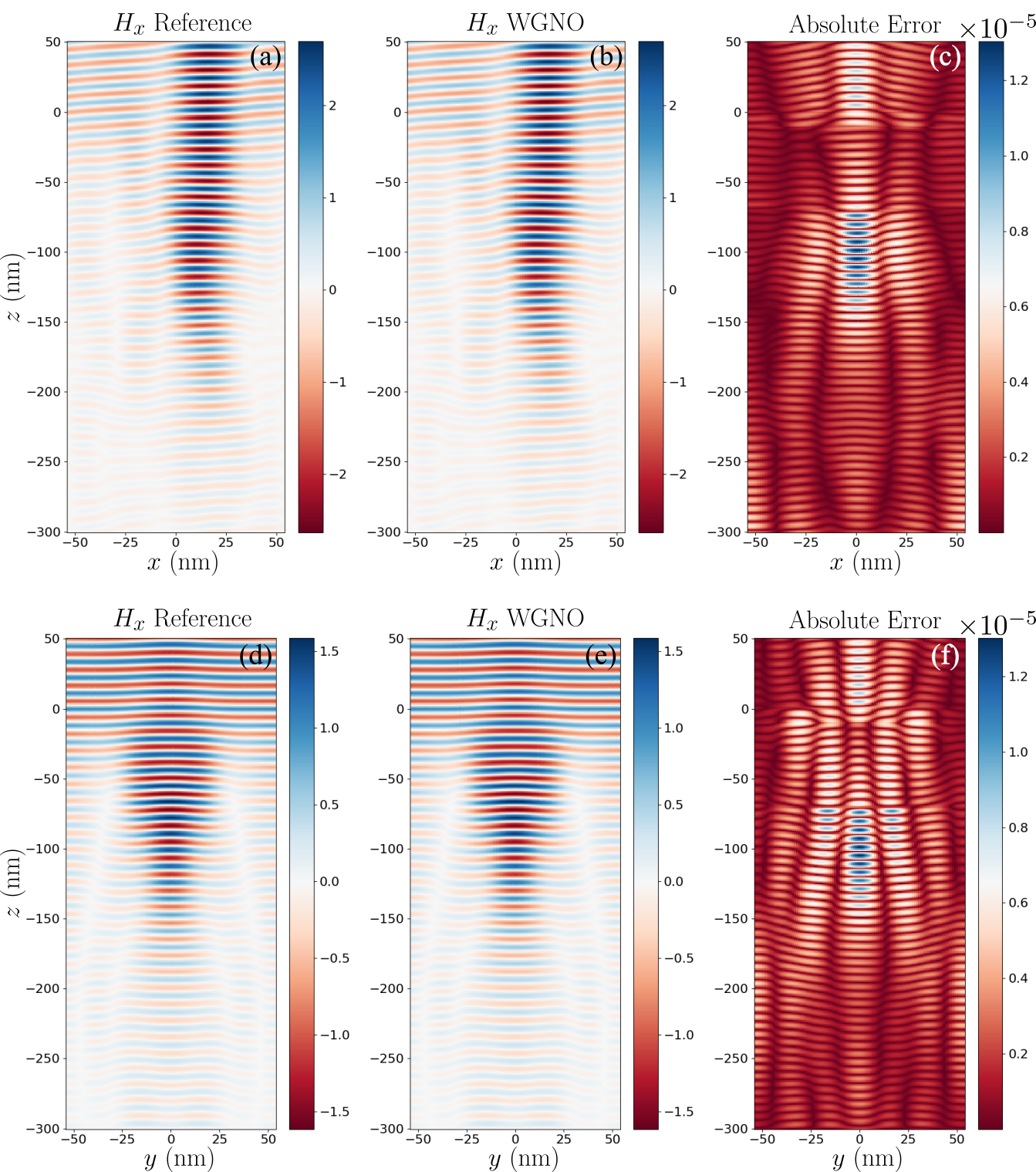}
	\caption{Comparison of reference WG solution ((a),(d)) and WGNO solution ((b),(e)) for a 3D mask at 11.2 nm, (c) and (f) are absolute errors. (a)--(c) are in cross-section $y=0$, (d)--(f) are in the cross-section $x=0$.}\label{fig3D_11_2nm}
\end{figure}

\begin{table}[h!]
	\centering
	\caption{Performance of WGNO on 3D lithography mask simulation. Training time less $20$ s.}
	\label{tab:3d_mask}
	\begin{tabular}{l|ccc}
		\toprule
		Wavelength & Rel. $L_2$ Error & Inference (s)& Training time (s) \\
		\midrule
		13.5 nm & $3.1\times 10^{-7}$ & $2.24\times 10^{-4}$ & 19.4 \\
		11.2 nm & $2.8\times 10^{-6}$ & $2.19\times 10^{-4}$ & 18.9 \\
		\bottomrule
	\end{tabular}
\end{table}

\subsubsection{Generalization}

For the evaluation of generalization properties of WGNO we use the measure of generalization $\mu$~(see in~\cite{ESKIN2025114085}) which is given in appendix~\ref{App3}. We investigate these properties of the described neural network using the 2D example~\ref{section2DWDNO} (the wavelength is 13.5 nm).

We used the following parameter of the considered problem $a \in [0.99 L_x / 4,L_x / 4]$, which is half the length of the hole in the absorbers. Consider how the generalization parameter of the solutions provided by WGNO depends on the number of partitions $N_q$ of the area of change of the parameter $a$. The calculation results are presented in the Table~\ref{genTab}. As can be seen from this table, the measure of generalizing properties $\mu$ increases with increasing size of $N_q$, and this measure of generalization quickly reaches values close to 1.

\begin{table}[h!]
	\centering
	\caption{Dependence of measure of generalization $\mu$ of WGNO on the number of partitions $N_q$ of the area of change of the parameter $a$.}
	\label{genTab}
	\begin{tabular}{l|c}
		\toprule
		$N_q$ & Measure of generalization $\mu$ \\
		\midrule
		5  & 0.94225 \\
		10 & 0.98473 \\
		20 & 0.99607 \\
		50 & 0.99922 \\
		100 & 0.99978 \\
		\bottomrule
	\end{tabular}
\end{table}

\section{Conclusion}

In this work, we have shown that physics-informed neural networks and neural operators can achieve high accuracy for complex diffraction problems of EUV lithography simulation. We determined that while training times can vary, the inference time is extremely small. For relatively thin masks, PINNs can provide solutions accurate enough for initial calculations. Most significantly, we have established that our proposed Waveguide neural operator is a state-of-the-art solver for this application. Using the example of solving a 2D problem, it is shown that the presented WGNO has pronounced generalizing properties: for problem parameters unseen during training, it provides a solution accuracy close to that of the parameters present in the training dataset. It provides highly accurate solutions for a full 3D mask, surpassing modern approaches in both accuracy and inference time, all while requiring only a very short training period. This level of performance makes the WGNO a highly promising tool for accelerating the OPC design cycle in semiconductor manufacturing. Furthermore, we are confident that the approaches we have developed are not limited to lithography and can be applied to study the properties and optimization of other complex electromagnetic structures, such as metamaterials and composite materials.

\newpage

\appendix

\section{Relative ${\mathbb L}_2$ error}\label{App1}
To evaluate the accuracy of the approximate solution obtained with the help of considered methods, the values of the solution of problems (\ref{eq2})--(\ref{eq4}) calculated by the numerical solver or predicted by the neural network at given points are compared with the exact solution or the values calculated on the basis of classical high-precision numerical methods. As a measure of accuracy, the relative total ${\mathbb L}_2$ error of prediction is taken, which can be expressed with the following relation
\begin{equation}\label{eqApp1}
	\epsilon_{\rm error} = \left\{\frac{1}{N_e} \sum^{N_e}_{i=1} \left[{\vec u}_{\pmb \theta}({x}_i,{y}_i) - {\vec u}({x}_i,{y}_i)\right]^2 \right\}^{1/2} \times \left\{\frac{1}{N_e} \sum^{N_e}_{i=1} \left[{\vec u}({x}_i,{y}_i)\right]^2 \right\}^{-1/2},
\end{equation}
where $\left\{{x}_i,{y}_i\right\}^{N_{e}}_{i=1}$ is the set of evaluation points taken from the domain, ${\vec u}_{\pmb \theta}$ and ${\vec u}$ are the predicted and reference solutions, respectively, $N_{e}$ is the number of evaluation points.

\section{Comparisons of the electric field distributions of the exact solution and the solutions given by the solvers}\label{App2}

Figures \ref{figA1}–\ref{figA4} present a comparison between the exact solutions of the test problems and those obtained using FreeMem++, the waveguide method, PINN, and WGNO. The selected test cases include: plane wave propagation in a homogeneous medium, reflection from a single interface (Figure~\ref{fig6} (a)), and reflection from a dielectric layer (Figure~\ref{fig6} (b)).

\begin{figure}[h]
	\centering
	\includegraphics[width=\linewidth]{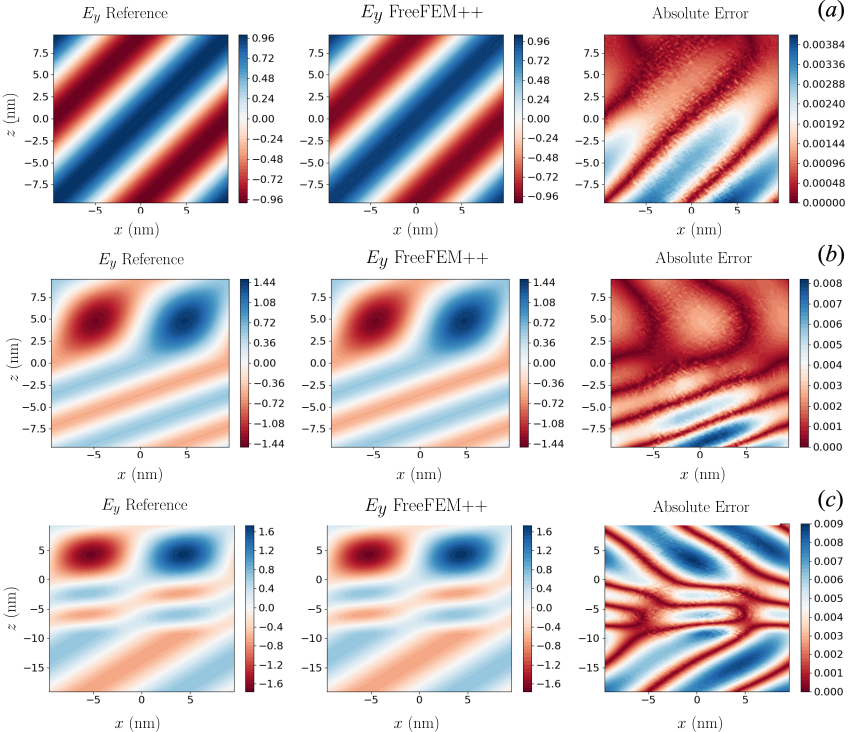}
	\caption{FreeFem++ solver. The left panels show the exact solutions, the middle panels show the solutions calculated using the given method, and the right panels show the absolute difference between the exact and calculated solutions. The relative error $\epsilon_{\rm error}$ are given in the table~\ref{tab:test_problems}. Panels (a), (b) and (c) correspond the problems of wave propagation in a homogeneous medium, reflection of an electromagnetic waves from a interface between two media and reflection from a dielectric layer, respectively.}
	\label{figA1}
\end{figure}

\begin{figure}[h]
	\centering
	\centering
	\includegraphics[width=\linewidth]{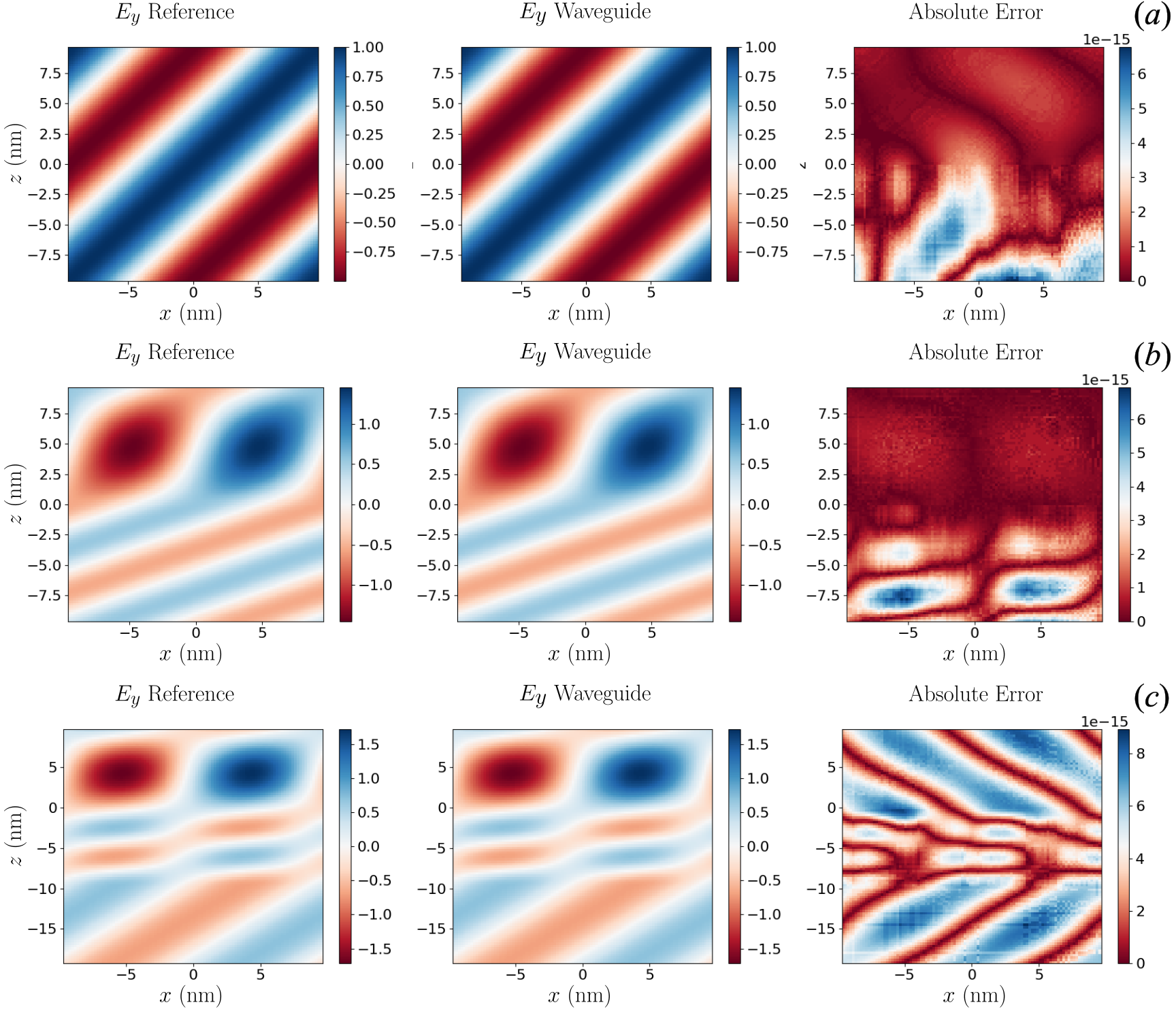}
	\caption{The same is shown in Figure~\ref{figA1} for the waveguide method.}
	\label{figA2}
\end{figure}

\begin{figure}[h]
	\centering
	\centering
	\includegraphics[width=\linewidth]{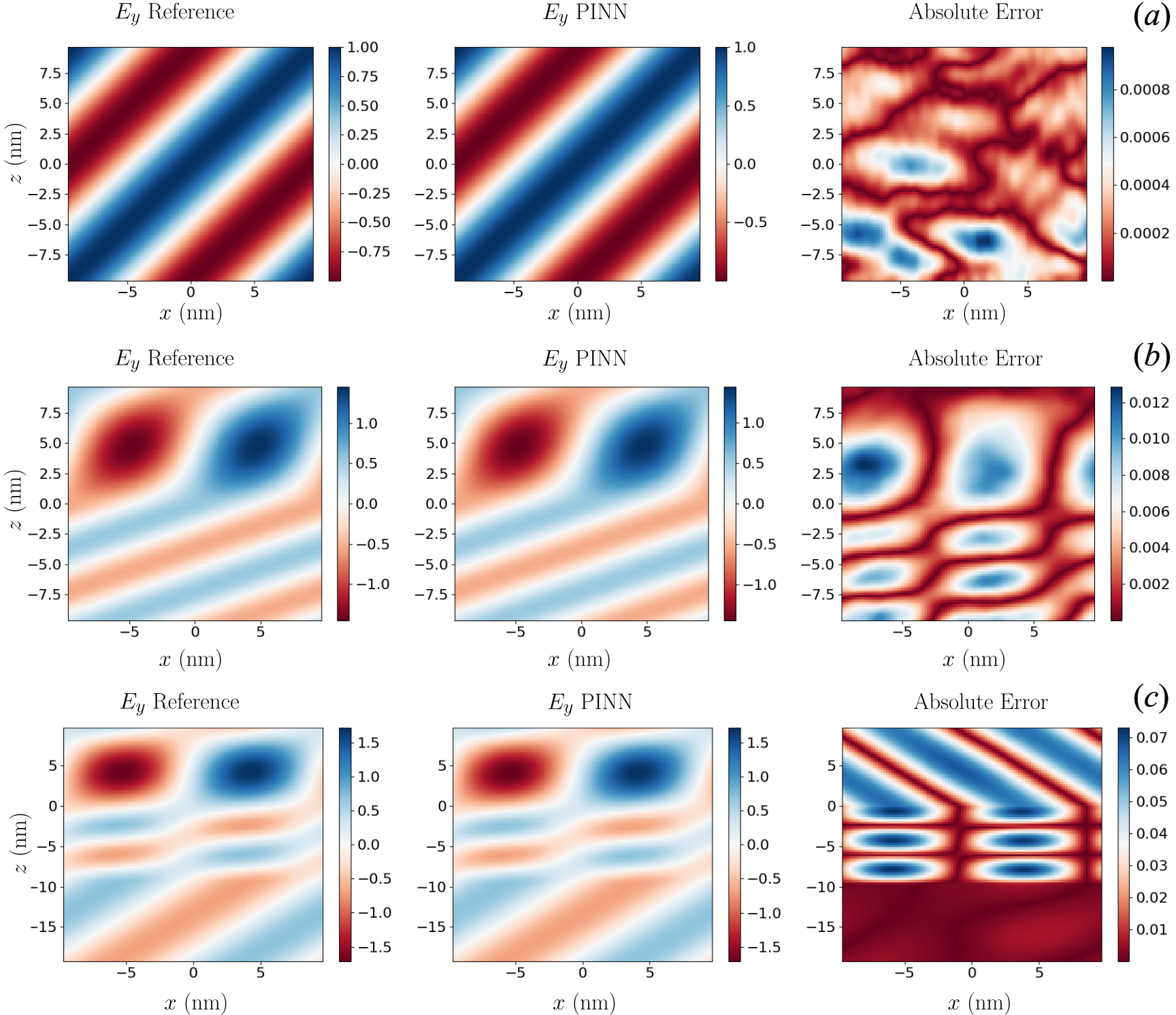}
	\caption{Approach based on the PINN. The left panels show the exact solutions, the middle panels show the calculated solutions using the given method, right panels are absolute difference between exact solutions and calculates solutions. The relative error $\epsilon_{\rm error}$ are given in the table~\ref{tab:test_problems}. Panels (a), (b) and (c) correspond the problems of wave propagation in a homogeneous medium, reflection of an electromagnetic waves from a interface between two media and reflection from a dielectric layer, respectively.}
	\label{figA3}
\end{figure}

\begin{figure}[h]
	\centering
	\includegraphics[width=\linewidth]{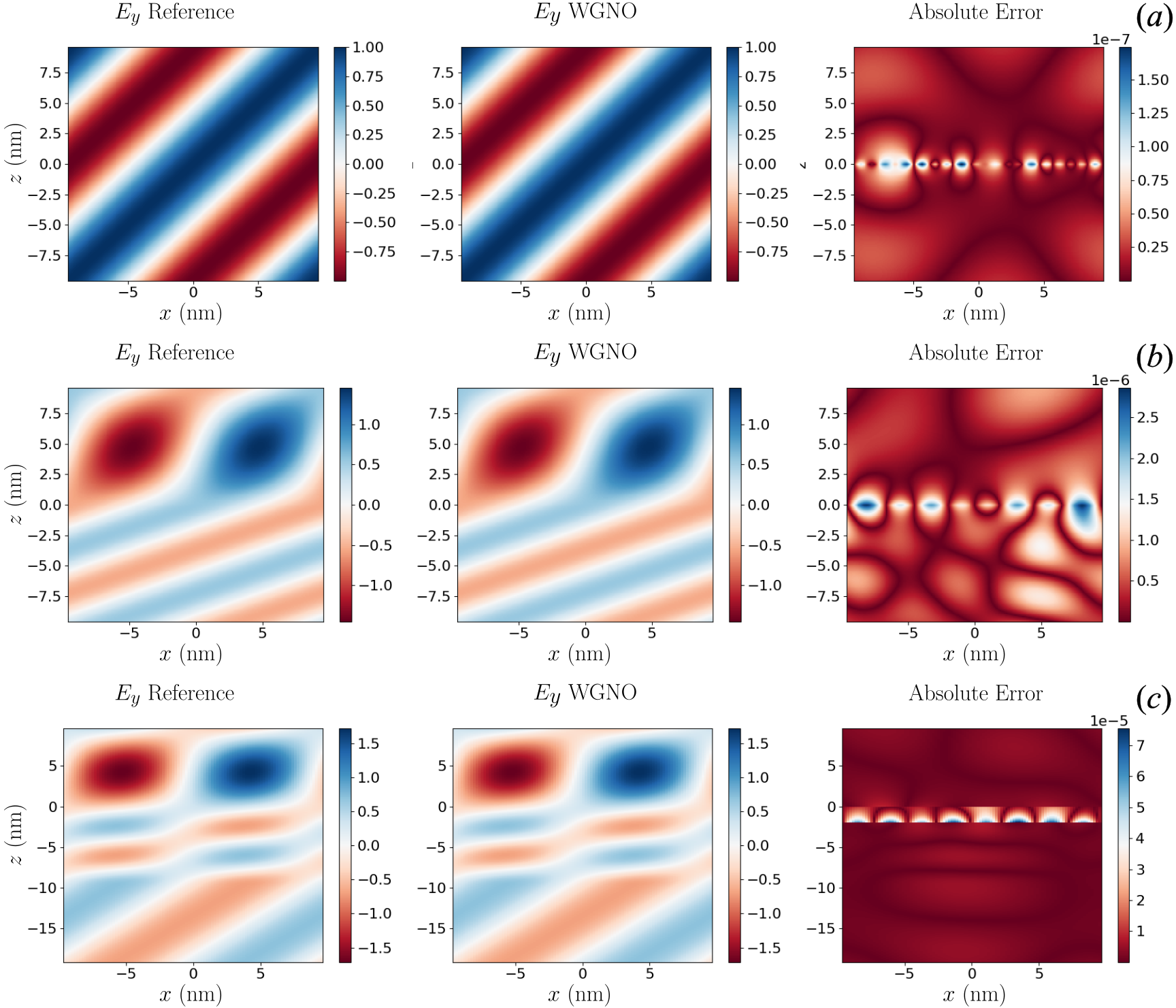}
	\caption{The same is shown in Figure~\ref{figA3} for the waveguide neural operator.}
	\label{figA4}
\end{figure}

\section{Measure of generalization properties}\label{App3}
To evaluate the generalization properties of the approximate solution obtained using the physics-informed methods, two sets of solutions predicted by the neural network on homogeneous grids of this parameter are required for the problem parameter $q$. The first set of solution values is calculated for the grid of the parameter $q$ in steps of $\Delta q$ for those values of $q$ at which the training was carried out. We denote this set of parameters as $\left\{{\vec q}_{i}\right\}^{N_{q}}_{i=1}$. The second set of solution values is calculated for the calculated grid of the parameter $q$ in steps of $\Delta q$, offset by $\Delta q/ 2$, i.e. at points as far away as possible from the nearest points of $q$ where the training was carried out, but not out of the training domain. We will denote this set of parameters as $\left\{{\vec q}_{i+1/2}\right\}^{N_{q}-1}_{i=1}$. As a measure of generalization properties, the ratio of the relative total ${\mathbb L}_2$ error of prediction of set $\left\{{\vec q}_{i+1/2}\right\}^{N_{q}-1}_{i=1}$ to the ratio of the relative total ${\mathbb L}_2$ error of prediction of set $\left\{{\vec q}_{i}\right\}^{N_{q}}_{i=1}$  is taken, which can be expressed with the following relation
\begin{align}\label{eq33}
	& \mu = \frac{\epsilon_1[{u}_{\bm \theta}, {u}]}{\epsilon_{1/2}[{u}_{\bm \theta}, {u}]}.
\end{align}
Here
\begin{align}
	& \epsilon_{1/2}[{u}_{\bm \theta}, {u}] = \left\{\sum^{N_q -1}_{j=1}\sum^{N_e}_{i=1} \left[{u}_{\bm \theta}({\vec x}_i,q_{j+1/2}) - {u}({\vec x}_i,q_{j+1/2})\right]^2 \right\}^{1/2} {\times} \left\{ \sum^{N_q -1}_{j=1}\sum^{N_e}_{i=1} \left[{u}({\vec x}_i,q_{j+1/2})\right]^2 \right\}^{-1/2},\notag\\
	&  \epsilon_{1}[{u}_{\bm \theta}, {u}] = \left\{\sum^{N_q}_{j=1}\sum^{N_e}_{i=1} \left[{u}_{\bm \theta}({\vec x}_i,q_{j}) - {u}({\vec x}_i,q_{j})\right]^2 \right\}^{1/2} {\times} \left\{ \sum^{N_q}_{j=1}\sum^{N_e}_{i=1} \left[{u}({\vec x}_i,q_{j})\right]^2 \right\}^{-1/2}, \notag
\end{align}
where $\left\{{\vec x}_{i}\right\}^{N_{e}}_{i=1}$ is the set of evaluation points taken in the computational domain, $\left\{{q}_{i}\right\}^{N_{q}}_{i=1}$ and $\left\{{q}_{i+1/2}\right\}^{N_{q}-1}_{i=1}$ are sets taken from domain $[q_{\rm min}, q_{\rm max}]$, ${u}_{\bm \theta}$ and ${u}$ are the predicted and reference solutions, respectively, the second argument of the functions ${u}_{\bm \theta}$ and ${u}$ denotes the parameter of the differential equation for which these functions were calculated.

Thus, the closer the value of the measure of generalization properties $\mu$ is to one, the neural network outputs solutions to the problem at the parameters of the problem at which it was not trained, closer in accuracy to its solutions to the problem at the parameters at which it was trained.

\newpage
\bibliographystyle{IEEEtran}
\bibliography{Eskin_2D_Mask}

\end{document}